\documentclass{article}[11]
\def\coralreport{0}
\def\ifproposal{1}

\usepackage[final]{nips_2016t}
\usepackage{comment}
\usepackage{subcaption}
\usepackage{graphicx}
\usepackage{fancyhdr, lipsum}
\usepackage{multicol}
\usepackage{float}
\usepackage{adjustbox}      
\usepackage{filecontents}
\newcommand{\add}[1]{{\color{black} {#1}}}

\usepackage[utf8]{inputenc} 
\usepackage[T1]{fontenc}    
\usepackage{hyperref}       
\usepackage{url}            
\usepackage{booktabs}       
\usepackage{amsfonts}       
\usepackage{nicefrac}       
\usepackage{microtype}      

\usepackage{natbib}
 \bibpunct[, ]{(}{)}{,}{a}{}{,}%
\usepackage[table,xcdraw]{xcolor}

\usepackage{caption}
\usepackage{graphicx}
\usepackage{amssymb}
\usepackage{amsmath,amsfonts,amsthm,graphicx}
\usepackage{enumerate}
\usepackage{verbatim}
\usepackage{multicol}
\usepackage[english]{babel}
\usepackage{fancyhdr, lipsum}
\usepackage{multicol}
\usepackage[table,xcdraw]{xcolor}
\usepackage{float}
\usepackage{color,soul}
\usepackage{multirow}
\usepackage[normalem]{ulem}
\usepackage{ifthen}
\newtheorem{prop}{Proposition}

\usepackage[colorinlistoftodos,bordercolor=orange,backgroundcolor=orange!20,linecolor=orange,textsize=scriptsize]{todonotes}
\newcommand{\argminF}{\mathop{\mathrm{argmin}}\limits}   

\ifthenelse{\coralreport=0}{
\title{Applying Deep Learning to the Newsvendor Problem}

\author{
	Afshin Oroojlooyjadid 
	\\
	Lehigh University\\
	Bethlehem, PA 18015 \\
	\texttt{oroojlooy@lehigh.edu} \\
	\And
	Lawrence V.~Snyder
	\\
	Lehigh University\\
	Bethlehem, PA 18015 \\
	\texttt{larry.snyder@lehigh.edu} \\
	\And
	Martin Tak\'a\v{c}
	\\
	Lehigh University\\
	Bethlehem, PA 18015 \\
	\texttt{takac.mt@gmail.com} \\
}
}

\begin{document}
\ifthenelse{\coralreport=1}{
\title{Applying Deep Learning to the Newsvendor Problem}

\ifthenelse{\coralreport=1}{
	\author{Afshin Oroojlooy, Lawrence Snyder, Martin Tak\'a\v{c}}
	\affil{Department of Industrial and Systems Engineering\\Lehigh University, Bethlehem, PA, USA}
	
	\titlepage
}{}
}
	\maketitle
	
	\vspace{1 in}
	\begin{abstract}
		The newsvendor problem is one of the most basic and widely applied inventory models. There are numerous extensions of this problem. 
		If the probability distribution of the demand is known, the problem can be solved analytically. 
		However, approximating the probability distribution is not easy and is prone to error; therefore, the resulting solution to the newsvendor problem may be not optimal. 
		To address this issue, we propose an algorithm based on deep learning that optimizes the order quantities for all products based on features of the demand data. Our algorithm integrates the forecasting and inventory-optimization steps, rather than solving them separately, as is typically done, and does not require  knowledge of the probability distributions of the demand. Numerical experiments on real-world data suggest that our algorithm outperforms other approaches, including data-driven and machine learning approaches, especially for demands with high volatility. Finally, in order to show how this approach can be used for other inventory optimization problems, we provide an extension for $(r,Q)$ policies.
	\end{abstract}
	
	\section{Introduction} \label{sec:Introduction} 
	
	The newsvendor problem optimizes the inventory of a perishable good. Perishable goods are those that have a limited selling season; they include fresh produce, newspapers, airline tickets, and fashion goods. The newsvendor problem assumes that the company purchases the goods at the beginning of  a time period and sells them during the period. 
	At the end of the period, unsold goods must be discarded, incurring a {\em holding cost}.
	In addition, if it runs out of the goods in the middle of the period, it incurs a {\em shortage cost}, losing potential profit.
	Therefore, the company wants to choose the order quantity that minimizes the expected sum of the two costs described above. The problem dates back to \cite{Edgeworth1888}; see \cite{Porteus08} for a history and \cite{Zi00}, \cite{Po02}, and \cite{snyder2018fundamentals}, among others, for textbook discussions.
	
	The optimal order quantity for the newsvendor problem can be obtained by solving the following optimization problem:
	\begin{equation} 
		\label{classicNWcost}
		\begin{array}{llll} 
			\min \limits_{y} C(y) = E_d \left[ c_p (d - y)^{+} + c_h (y - d)^{+} \right],
		\end{array}
	\end{equation} 
	where $d$ is the random demand, $y$ is the order quantity, $c_p$ and $c_h$ are the per-unit shortage and holding costs (respectively), and $(a)^{+} := \max\{0,a\}$.
	In the classical version of the problem, the shape of the demand distribution (e.g., normal) is known, and the distribution parameters are either known or estimated using available (training) data. If $F(\cdot)$ is the cumulative density function of the demand distribution and $F^{-1}(\cdot)$ is its inverse, then the optimal solution of \eqref{classicNWcost} can be obtained as
	\begin{equation}
		\label{NW_optimal_solution}
		y^\ast = F^{-1}\left(\frac{c_p}{c_p+c_h}\right) = F^{-1}(\alpha),
	\end{equation}
where $\alpha = c_p/(c_p+c_h)$ 
(see, e.g., \cite{snyder2018fundamentals}).
	
	Extensions of the newsvendor problem are too numerous to enumerate here (see \cite{Choi2012} for examples); instead, we mention two extensions that are relevant to our model. First, in real-world problems, companies rarely manage only a single item, so it is important for the model to provide solutions for multiple items. 
	(We do not consider substitution, demand correlation, and complementarity effects as \cite{bassok1999single, nagarajan2008inventory} do for the multi-product newsvendor problem.)
	Second, companies often have access to some additional data---called {\em features}---along with the demand information. These might include weather conditions, day of the week, month of the year, store location, etc \cite{vahn_News}. The goal is to choose today's base-stock level, given the observation of today's features. 
	We will call this problem multi-feature newsvendor (MFNV) problem.
	In this paper, we propose an approach for solving this problem that is based on deep learning, i.e., deep neural networks (DNN).
	
	The remainder of this paper is structured as follows. A brief summary of the literature relevant to the MFNV problem is presented in Section \ref{sec:litreview}. 
	Section \ref{sec:methods} presents the details of the proposed algorithm. 
	Numerical experiments are provided in Section \ref{sec:Numerical Experiments}. Section \ref{sec:r_q} introduces an extension of the approach for $(r,Q)$ policies, and the conclusion and a discussion of future research complete the paper in Section \ref{sec:conclusion}.
	
	\section{Literature Review}\label{sec:litreview}

	\subsection{Current State of the Art} \label{sec:stateofart}
\begin{figure}
		\centering
		\begin{subfigure}{0.35\textwidth}
			\centering
			\includegraphics[width=0.65\textwidth]{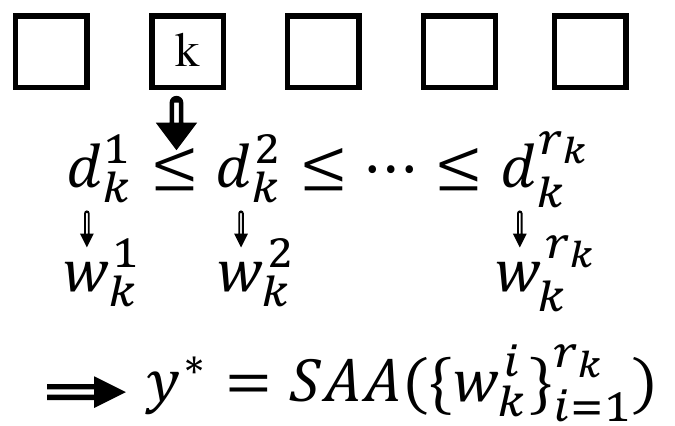}
			\caption{K-nearest neighbors (KNN) and random forest (RF).}
			\label{fig:mfnv_saa}
		\end{subfigure} %
		\begin{subfigure}{0.35\textwidth}
			\centering
			\includegraphics[width=0.65\textwidth]{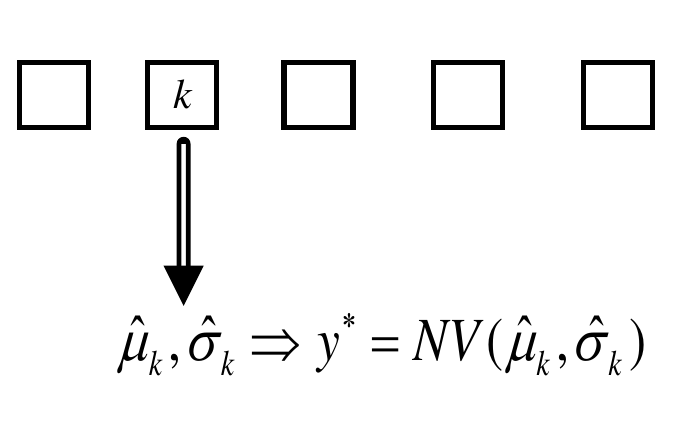}
			\caption{Separated estimation and optimization (SEO).}
			\label{fig:mfnv_seo}
		\end{subfigure}
		
		\begin{subfigure}{0.33\textwidth}
			\centering
			\includegraphics[width=0.65\textwidth]{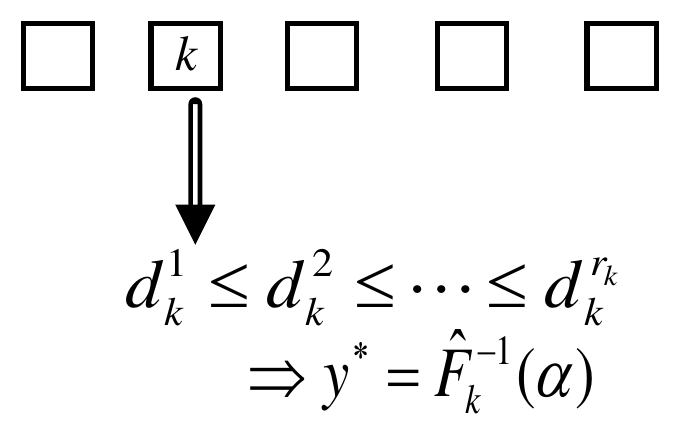}
			\caption{Empirical quantile (EQ).}
			\label{fig:mfnv_eq}
		\end{subfigure} %
		\begin{subfigure}{0.33\textwidth}
			\centering
			\includegraphics[width=0.65\textwidth]{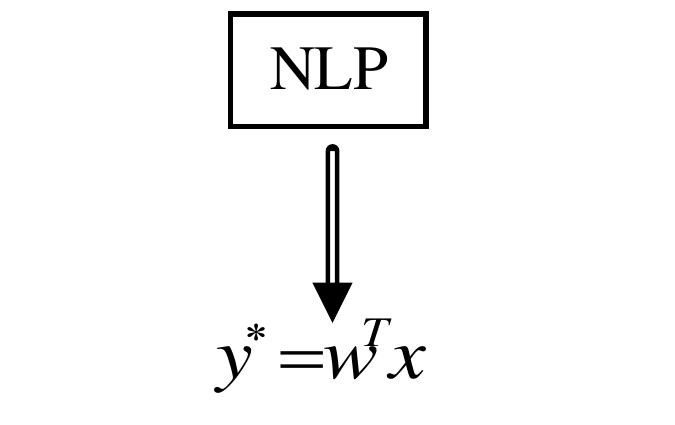}
			\caption{Linear machine learning (LML).}
			\label{fig:mfnv_lml}
		\end{subfigure} %
		\begin{subfigure}{0.33\textwidth}
			\centering
			\includegraphics[width=0.65\textwidth]{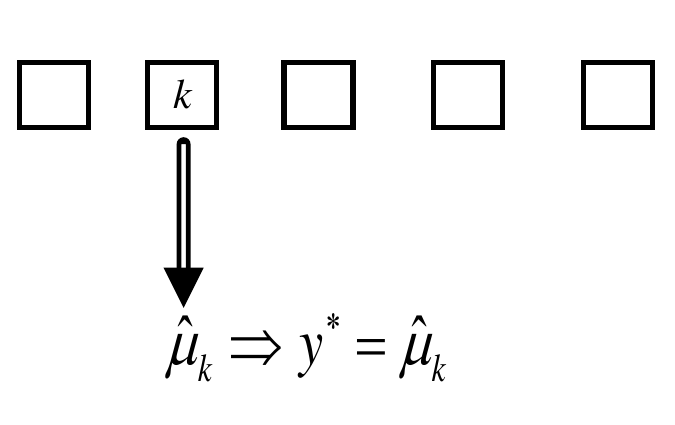}
			\caption{Estimate-as-solution (EAS).}
			\label{fig:mfnv_eas}
		\end{subfigure} %
		\caption{Approaches for solving MFNV problem. Squares represent clusters.}
		\label{fig:mfnv}
	\end{figure}

	Currently, there are five main approaches in the literature for solving MFNV. 
	The first category, which we will call the {\em estimate-as-solution} (EAS) approach, involves first clustering the demand observations, then forecasting the demand, and then simply treating the point forecast as a deterministic demand value, i.e., setting the newsvendor solution equal to the forecast. (See Figure~\ref{fig:mfnv_eas}, which shows cluster $k$ and the order quantity, which is simply the forecast.) 
	By clustering, we mean that all demand observation that have same feature values are put together in a set, called a cluster. For example, when there are 100 demand records for two products in two stores, there are four clusters, and on average each cluster has 25 records.
	The forecast may be performed in a number of ways, some of which we review in the next few paragraphs.		
	
	This approach ignores the key insight from the newsvendor problem, namely, that we should not simply order up to the mean demand, but rather choose a level that strikes a balance between underage and overage costs using the distribution of the demand.
	Nevertheless, the approach is common in the literature.
	For example, \cite{yu2013support} propose a support vector machine (SVM) model to forecast newspaper demands at different types of stores, along with 32 other features. 
	\cite{wu2011support} use a weighted support vector regression (SVR) model to forecast warranty claims; 
	their model gives more priority to the most recent warranty claims.	
	\cite{chi2007modeling} propose a SVM model to determine the replenishment point in a vendor-managed replenishment system, and a genetic algorithm is used to solve it. 
	\cite{carbonneau2008application} present a least squares SVM (LS-SVM) model to forecast a manufacturer's demand. 
	\cite{ali2013selecting} forecast grocery sales, with datasets containing millions of records, and for each record there are thousands of features. They reduce the number of features and data and use SVM to solve the problem.
	\cite{lu2014hybrid} propose an iterative algorithm to predict sales. 
	They propose an algorithm based on independent component analysis, $k$-mean clustering, and SVR to provide the prediction. 
	
	Classical parametric approaches for forecasting include ARIMA, TRANSFER, and GARCH models \citep{box2015time,shumway2010time}; these are also used for demand forecasting (see \citet{cardoso2007newspaper,shukla2011arima}).
	Similarly, \cite{taylor2000quantile} uses a normal distribution to forecast demand one or more time steps ahead; however, his model does not perform well when demands are correlated over time and when the demands are volatile. 
	These and other limitations have motivated the use of DNN  to obtain demand forecasts. For example,
	\cite{efendigil2009decision} propose a DNN model to forecast demand based on recent sales, promotions, product quality, and so on. 
	\cite{DBLPjournalsCorrVieira15} propose a deep learning algorithm to predict online activity patterns that result in an online purchase. 
	For reviews of the use of DNN for forecasting, see \cite{ko2010review,kourentzes2010advances,qiu2014ensemble, crone2011advances}.
	
	The common theme in all of the papers in the last two paragraphs is that they provide only a forecast of the demand, which must then be treated as the solution to the MFNV or other optimization problem. This is the EAS approach.
	
	The second approach for solving MFNV-type problems, which \cite{vahn_News} refer to as {\em separated estimation and optimization} (SEO), involves first estimating (forecasting) the demand distribution and then plugging the estimate into an optimization problem such as the classical newsvendor problem. 
	The estimation step is performed similarly as in the EAS approach except that we estimate more than just the mean. 
	For example, we might estimate both mean ($\mu_k$) and standard deviation ($\sigma_k$) for each cluster, which we can then use in the optimization step. 
	(See Figure~\ref{fig:mfnv_seo}.) Or we might use the $\sigma$ that was assumed for the error term in a regression model. 
	The main disadvantage of this approach is that it requires us to assume a particular form of the demand distribution (e.g., normal), whereas empirical demand distributions are often unknown or do not follow a regular form. 
	A secondary issue is that we compound the data-estimation error with model-optimality error. 
	\cite{vahn_News} show that for some realistic settings, the SEO approach is provably suboptimal. 
	This idea is used widely in practice and in the literature; a broad list of research that uses this approach is given by \cite{turken2012multi}.  \cite{vahn_News} analyze it as a straw-man against which to compare their solution approach.

	The third approach was proposed by \cite{Thiele} for the classical newsvendor problem. 
	Their approach involves sorting the demand observations in ascending order $d_1 \le d_2 \le \cdots \le d_n$ and then estimating the $\alpha$th quantile of the demand distribution, $F^{-1}(\alpha)$, using the observation that falls $100\alpha$\% of the way through the sorted list, i.e., it selects the demand $d_j$ such that $j = \lceil n \frac{c_p}{c_p + c_h}\rceil$.
	This quantile is then used as the base-stock level, in light of \eqref{NW_optimal_solution}.
	Since they approximate the $\alpha$th quantile, we refer to their method as the {\em empirical quantile} (EQ) method. (See Figure~\ref{fig:mfnv_eq}.) 
	Importantly, EQ does not assume a particular form of the demand distribution and does not approximate the probability distribution, so it avoids those pitfalls. 
	However, an important shortcoming of this approach is that it does not use the information from features.
	In principle, one could extend their approach to the MFNV by first clustering the demand observations and then applying their method to each cluster. 
	However, similar to the classical newsvendor algorithm, this would only allow it to consider categorical features and not continuous features, which are common in supply chain demand data, e.g. \cite{ali2013selecting} and \cite{vahn_News}. 
	Moreover, even if we use this clustering approach, the method cannot utilize any knowledge from other data clusters, which contain valuable information that can be useful for all other clusters. 
	Finally, when there is volatility among the training data, the estimated quantile may not be sufficiently accurate, and the accuracy of EQ approach tends to be worse. 
	
	In the newsvendor problem, the optimal solution is a given quantile of the demand distribution. 
	Thus, the problem can be modeled as a quantile regression problem, in a manner similar to the empirical quantile model of \cite{Thiele}.
	\citet{taylor2000quantile} was the first to propose the use of neural networks as a nonlinear approximator of the quantile regression to get a conditional
	density of multi-period financial returns. 
	Subsequently, several papers used quantile-regression neural networks to obtain a quantile regression value. For example, 
	\cite{cannon2011quantile} uses a quantile-regression neural network to predict daily precipitation; \cite{el2014quantile} uses it to predict drug activities; and 
	\cite{xu2016quantile} uses a quantile autoregression neural network to evaluate value-at-risk. 
	One can consider our approach as a quantile-regression neural network for the newsvendor problem. 
	However, our approach is much more general and can be applied to other inventory optimization problems, provided that a closed-form cost function exists. To demonstrate this, in Section \ref{sec:r_q} we extend our approach to solve an inventory problem that does not have a quantile-type solution, namely, optimizing the parameters of a $(r,Q)$ policy.

	A fourth approach for solving MFNV-type problems can be derived from the method proposed by \cite{bertsimas2014predictive}, which applies several machine learning (ML) methods on a general optimization problem given by
	\begin{equation} z^\ast(x)  = \argminF_{z} \mathbb{E} \left[ c(z,y) | x\right], \label{eq:bertsimas_obj} \end{equation}
	where $\left\{(x^1, y^1), \dots, (x^N, y^N)\right\}$ are the available data---in particular, $x^i$ is a $d$-dimensional vector of feature values and $y^i$ is the uncertain quantity of interest, e.g., demand values---and $z$ is the decision variable. 
	They test five algorithms to optimize \eqref{eq:bertsimas_obj}: $k$-nearest neighbor (KNN), random forest (RF), kernel method, classification and regression trees (CART), and locally weighted scatterplot smoothing (LOESS). They use sample average approximation (SAA) as a baseline, and each algorithm provides substitute weights for the SAA method. 
	For example, KNN identifies the set of $k$ nearest historical records to the new observation $x$ such that
	$$ \mathcal{N}(x) = \left \{ i=1,\dots,n : \sum_{j=1}^{n} \mathbb{I} \{||x-x_i|| \ge ||x-x_j|| \} \le k \right \}.$$
	\cite{bertsimas2014predictive} assign weights $w_i = 1 / k$ for all $i \in \mathcal{N}(x)$ (and zero otherwise) and call a weighted SAA; for example, if applied to the newsvendor problem, the SAA might take the form
	\begin{equation}
	\label{eq:weighted_SAA}
	q=\inf \left \{ d_j : \sum_{i=1}^{j} w_i \ge \frac{c_p}{c_p + c_h} \right \},
	\end{equation}
	where $d_j$ are the ascending sorted demands (see Figure \ref{fig:mfnv_saa}). Similarly, in RF, there are $T$ trees. The weight of each observation is obtained using 
	$$w_i = \frac1T \sum_{t=1}^{T} \frac{\mathbb{I} \{ R^t(x) = R^t(x_i) \} }{| \{j : R^t(x_j) = R^t(x^i)\}|},$$
	where $R^t(x)$ is the region of tree $t$ that observation $x$ is in. 
	In other words, the RF algorithm counts all trees in which the new observation $x$ is in the same region as historical observation $x_i$, $i=1,\dots, n$, and normalizes them over all observations in tree $t$ that have the same region. Finally, it normalizes the weights over all trees. Using these weights, the method of \cite{bertsimas2014predictive} as applied to the newsvendor problem calls the weighted SAA \eqref{eq:weighted_SAA} to get the order quantity. 
	\cite{bertsimas2014predictive} discuss asymptotic convergence of their methods and compare their performance with that of SAA. 
	
	The fifth approach for the MFNV, and the one that is closest to our proposed approach, was introduced by \cite{vahn_News}; we refer to it as the {\em linear machine learning} (LML) method. 
	They postulate that the optimal base-stock level is related to the demand features via a linear function; that is, that $y^* = w^Tx$, 
	where $x$ is the vector of features and $w$ is a vector of (unknown) weights.
	
	They estimate these weights by solving the following nonlinear optimization problem, essentially fitting the solution using the newsvendor cost:
	\begin{equation}
	\label{eq:vahn2}
	\begin{array}{l}
		\textstyle{\min_{w}}\ \frac{1}{n} \sum_{i=1}^{n} \left[c_p(d_i - w^Tx_i)^{+} + c_h(w^Tx_i - d_i)^{+} \right] + \lambda||w||^2_k \\
		\text{s.t.}\ (d_i - w^Tx_i)^{+} \ge d_i - w_1 - \sum \limits_{j=2}^p w_i x_i^j ; ~~ \forall i=1, \dots, n\\
		\phantom{\text{s.t.}\ }(w^Tx_i - d_i)^{+} \ge w_1 + \sum \limits_{j=2}^p w_i x_i^j - d_i ; ~~ \forall i=1, \dots, n\\				
	\end{array}
	\end{equation}
	where $n$ is the number of observations, $p$ is the number of features, and $\lambda||w||^2_k$ is a regularization term. 
	The LML method avoids having to cluster the data, as well as having to specify the form of the demand distribution. \cite{vahn_News} comprehensively analyze the effects of adding nonlinear combination of features into the feature space, as well as the effects of regularization and of overfitting. (For more theoretical details on these concepts, see \cite{smola2004tutorial}.)
	However, this model does not work well when $p \gg n$ and its learning is limited to the current training data. 
	In addition, if the training data contains only a small number of observations for each combination of the features, the model learns poorly. 
 	Finally, it makes the strong assumption that $x$ and $y^*$ have a linear relationship. 
	We drop this assumption in our model and instead use DNN to quantify the relationship between $x$ and $y^*$; see Section \ref{sec:methods}.
	\cite{vahn_News} also propose a kernel regression (KR) model to optimize the order quantity, in which weighted historical demands are used to build an empirical cdf of the demand. The weights are proportional to the distance of the newly observed feature value with historical feature values, i.e.,
	$$w_i = \frac{K ( x - x_i) }{\sum_{j=1}^{n} K( x - x_j) },$$
where $K(u) = \exp (-||u||_2^2/2h)/\sqrt{2\pi}$ and $h$ is the kernel bandwidth that has to be tuned. 
	Then they call weighted SAA \eqref{eq:weighted_SAA} to obtain the order quantity. 
	In addition, they provide a mathematical analysis of the generalization errors associated with each method.

	There is a large body of literature on data-driven inventory management that assumes we do not know the demand distribution and instead must directly use the data to make a decision.	
	\cite{besbes2013implications} consider censored data (in which some demands cannot be observed due to stockouts) in the newsvendor problem. 
	The paper proposes three models and algorithms to minimize the regret when real, censored, and partially censored demand are available. 
They propose an EQ-type algorithm (discussed above) for observable demand. For censored and partially censored demand, they propose two algorithms, as well as lower and upper bounds on the regret value for all algorithms.
	\cite{burnetas2000adaptive} propose an adaptive model to optimize price and order quantity for perishable products with an unknown demand distribution, assuming historical data of censored sales are available. 
	They assume that the demand is continuous 
	and propose two algorithms, one for a fixed price and another for the pricing/ordering problem. 
	Their algorithm for choosing the order quantity provides an adaptive policy and works even when there is nearly no historical information, so it is suitable for new products. 
	It starts from an arbitrary point $q_0$ and iteratively updates it with some learning rate and information about whether or not the order quantity $q_t$ was sufficient to satisfy the demand in period $t$. 	
	
	None of these papers use features, which is the key aspect of our model. 
	One data-driven approach that does use features is by \cite{ban2017dynamic}, who propose a model to choose the order quantity for new, short-life-cycle products from multiple suppliers over a finite time horizon, assuming that each demand has some feature information. 
	They propose a data-driven algorithm, called the residual tree method, which is an extension of the scenario tree method from stochastic programming, and prove that this method is asymptotically optimal as the size of the data set grows. Their approach has separate steps for estimation (using regression) and optimization (using stochastic linear programming). Although their problem has some similarities to ours, it is not immediately applicable since it is designed for finite-horizon problems with multiple suppliers.  

	\subsection{Deep Learning}

	In this paper, we develop a new approach to solve the newsvendor problem with data features, based on deep learning. 
	Deep learning, or deep neural networks (DNN), is a branch of machine learning that aims to build a model between inputs and outputs. Deep learning has many applications in image processing, speech recognition, drug and genomics discovery, time series forecasting, weather prediction, and---most relevant to our work---demand prediction.
	On the other hand, one major criticism of deep learning (in non-vision-based tasks) is that it lacks interpretability---that is, it is hard for a user to discern a relationship between model inputs and outputs; see, e.g. \cite{lipton2016mythos}. In addition, it usually needs careful hyper-parameter tuning, and the training process can take many hours or even days. 
	We provide only a brief overview of deep learning here; for comprehensive reviews of the algorithm and its applications, see \cite{goodfellow2016deep}, \cite{schmidhuber2015deep}, \cite{lecun2015deep}, \cite{deng2013recent}, \cite{QiuDeepForecast2014}, \cite{NIPS2015_5955}, 
	and \cite{Langkvist201411}.
	
\begin{figure}[]
		\centering
		\caption{A simple deep neural network.}
		\label{dnn}
		\includegraphics[scale=0.3]{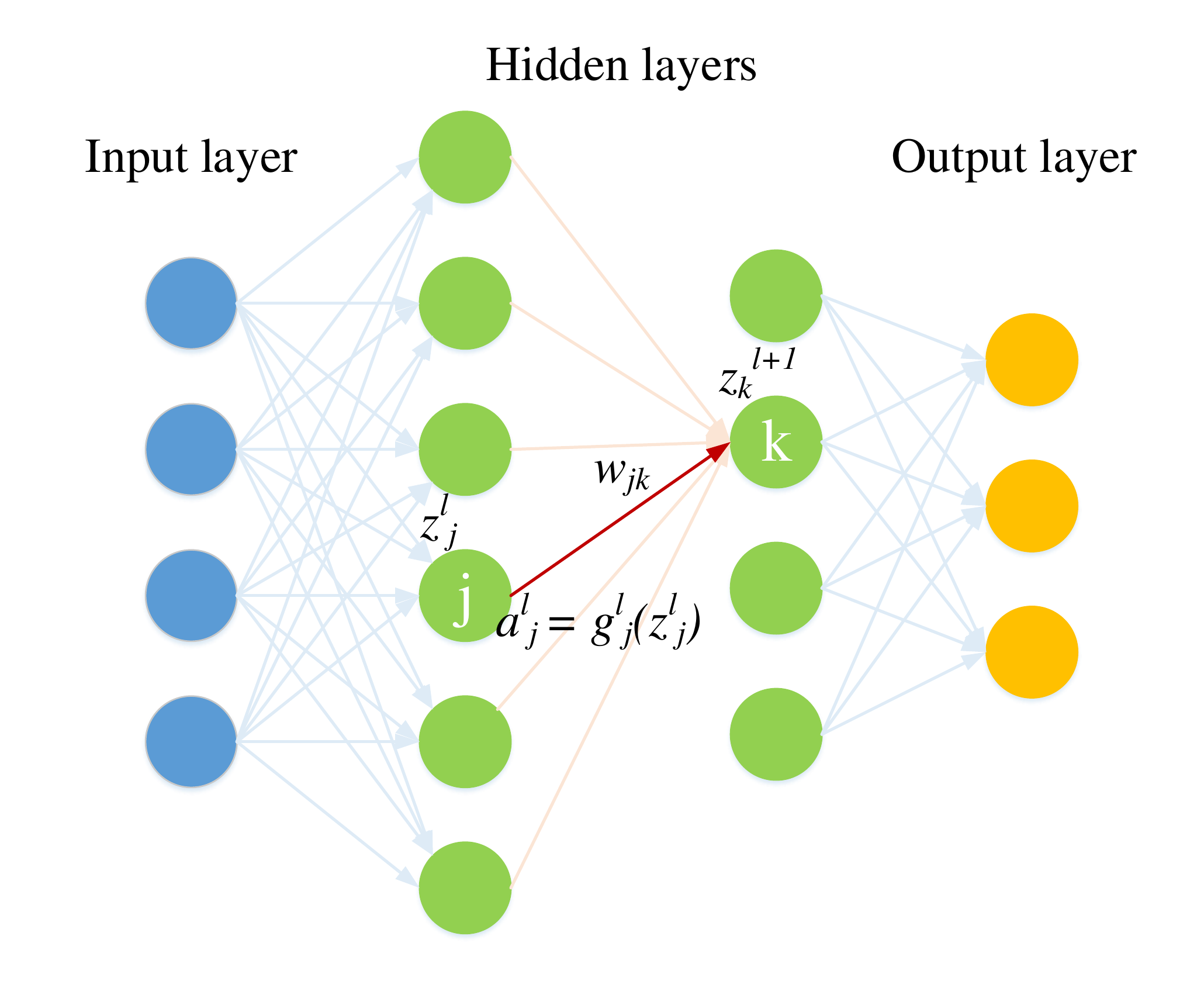}
	\end{figure}

	DNN uses a cascade of many layers of linear or nonlinear functions to obtain the output values from inputs. 
	A general view of a DNN is shown in Figure \ref{dnn}.
	The goal is to determine the weights of the network such that a given set of inputs results in a true set of outputs. A loss function is used to measure the closeness of the outputs of the model and the true values. 
	The most common loss functions are the hinge, logistic regression, softmax, and Euclidean loss functions.
	The goal of the network is to provide a small loss value, i.e., to optimize:
	\begin{equation*}
		\min_{w} \frac1n \sum_{i=1}^n \ell( \theta(x_i;w), y_i)   + \lambda R(w),
	\end{equation*}
	where $w$ is the matrix of the weights, $x_i$ is the vector of the inputs from the $i$th instance, $\theta(.)$ is the DNN function, and $R(w)$ is a regularization function with weight $\lambda$. The regularization term prevents over-fitting and is typically the $\ell_1$ or $\ell_2$ norm of the weights. (Over-fitting means that the model learns to do well on the training set but does not extend to the out-of-training samples; this is to be avoided.) 
	
	In each node $j$ ($j=1,\ldots,n$) of a layer $l$ ($l=1,\ldots,L$), the input value 
	\begin{equation}
		z_j^l =\sum \limits_{i=1}^n a_{i}^{l-1} w_{i,j}
	\label{eq:zjl}
	\end{equation}
	is calculated and the value of the function $g^l_j (z_j^l)$ provides the output value of the node. 
	The function $g^l_j(\cdot)$ is called the activation function; the value of $g^l_j (z_j^l )$ is called the activation of the node, and is denoted by $a^l_j$.
	Typically, all nodes in the network have similar $g^l_j(\cdot)$ functions. The most commonly used activation functions are the sigmoid ($1/(1+e^{-z_j^l})$) and tanh ($(1-e^{-2z_j^l})/(1+e^{-2z_j^l})$) functions, which add non-linearity into the model (see more details about them in \cite{lecun2015deep,goodfellow2016deep}).
	The activation function value of each node is the input for the next layer, and finally, the activation function values of the nodes in the last layer determine the output values of the network. The general flow of the calculations between two layers of the DNN, with a focus on $z_j^l$, $a_j^l$, $w_{jk}$, and $z_j^{l+1}$, is shown in Figure \ref{dnn}.
	
	In each DNN, the number of layers, the number of nodes in each layer, the activation function inside each node, and the loss function have to be determined. 
	After selecting those characteristics and building the network, DNN starts with some random initial solution. 
	In each iteration, the activation values and the loss function are calculated. Then, the back-propagation algorithm obtains the gradient of the network and, using one of several optimization algorithms \citep{rumelhart1988learning}, the new weights are determined. The most common optimization algorithms are gradient descent, stochastic gradient descent (SGD), SGD with momentum, and Adam optimizer (for details on each optimization algorithm see \cite{goodfellow2016deep}).
	This procedure is performed iteratively until some stopping condition is reached; typical stopping conditions are (a) reaching a maximum number of iterations and (b) attaining $ || \nabla_w \ell( \theta(x_i;w), y_i) || \le \epsilon $ through the back-propagation algorithm.

	Since the number of instances, i.e., the number of training records, is typically large, it is common \citep{goodfellow2016deep, bottou2010large} to use a stochastic approximation of the objective function. 
	That is, in each iteration, a mini-batch of the instances is selected and the objective is calculated only for those instances. 
	This approximation does not affect the provable convergence of the method. For example, in networks with sigmoid activation functions in which a quadratic loss function is used, the loss function asymptotically converges to zero if either gradient descent or stochastic gradient descent are used \citep{tesauro1989asymptotic, bottou2010large}.

	\subsection{Our Contribution}	
	To adapt the deep learning algorithm for the newsvendor problem with data features, we propose a revised loss function, which considers the impact of inventory shortage and holding costs. 
	The revised loss function \emph{allows the deep learning algorithm to obtain the minimizer of the newsvendor cost function directly, rather than first estimating the demand distribution and then choosing an order quantity.}

	In the presence of sufficient historical data, this approach can solve problems with known probability distributions  as accurately as \eqref{NW_optimal_solution} solves them. However, the real value of our approach is that it is effective for problems with small quantities of historical data, problems with unknown/unfitted probability distributions, or problems with volatile historical data---all cases for which the current approaches fail.
		
	\section{Deep Learning Algorithm for Newsvendor with Data Features} \label{sec:methods} 

	In this section, we present the details of our approach for solving the newsvendor problem with data features. 
	Assume there are $n$ historical demand observations for $m$ products. Also, for each demand observation, the values of $p$ features are known. That is, the data can be represented as
	$$\left\{(x^1_i,d^1_i),\dots, (x^m_i, d^m_i)\right\}_{i=1}^n,$$
where $x^q_i \in {\mathbb R}^p$ and $d^q_i \in {\mathbb R}$ for $i=1,\dots, n$ and $q = 1 , \dots , m$.
	The problem is formulated mathematically in \eqref{multi-newsvendor} for a given period $i$, $i=1,\ldots,n$, resulting in the order quantities $y_i^1, \dots , y_i^m$: 
	\begin{equation}
		\label{multi-newsvendor}
		E_i = \min\limits_{y_i^1 , \dots , y_i^m} ~ \frac{1}{m} \left[ \sum_{q=1}^{m} c_h (y^q_i - d^q_i)^{+} + c_p (d^q_i - y^q_i)^{+} \right], 
	\end{equation}
	where $E_i$ is the loss value of period $i$ and $E = \frac1n \sum_{i=1}^{n} E_i$ is the total loss value. 
	Since at least one of the two terms in each term of the sum must be zero, the loss function \eqref{multi-newsvendor} can be written as: 
	\begin{equation} 
	\label{original_nw}
	\begin{split}
	E_i & = \sum_{q=1}^m E_i^q \\
	E_i^q & = \begin{cases} c_p (d^q_i - y^q_i) ~, & \text{if}~y^q_i < d^q_i, \\
	c_h (y^q_i - d^q_i) ~, & \text{if}~d^q_i \le y^q_i. \end{cases}
	\end{split}
	\end{equation} 
	%
	%
	As noted above, there are many studies on the application of deep learning for demand prediction (see \cite{NIPS2015_5955}).
	Most of this research uses the Euclidean loss function
	(see \cite{qiu2014ensemble}).
	However, the demand forecast is an estimate of the first moment of the demand probability distribution; it is not, however, the optimal solution of model \eqref{multi-newsvendor}. 
	Therefore, another optimization problem must be solved to translate the demand forecasts into a set of order quantities. 
	This is the separated estimation and optimization (SEO) approach described in Section~\ref{sec:stateofart}, which may result in a non-optimal order quantity (\cite{vahn_News}).
	To address this issue, we propose two loss functions, the newsvendor cost function \eqref{multi-newsvendor} and a revised Euclidean loss function, so that instead of simply predicting the demand, the DNN minimizes the newsvendor cost function.
	Thus, running the corresponding deep learning algorithm gives the order quantity directly.
	
	We found that squaring the cost for each product\ in \eqref{multi-newsvendor} sometimes leads to better solutions, since the function is smooth, and the gradient is available in the whole solution space. Therefore, we also test the following revised Euclidean loss function:
	\begin{equation} 
	\label{quadratic_nw_loss}
		E_i = \min\limits_{y_1 , \cdots , y_n} ~ \frac{1}{m}
	\left[\sum_{i=1}^{m} \left[c_p (d^q_i - y^q_i)^+ + c_h (y^q_i - d^q_i)^+ \right]^2\right], \forall i=1,\dots,n
	\end{equation}
	which penalizes the order quantities that are far from $d_i$ much more than those that are close.  
	%
	Then we have
	\begin{equation} 
		\label{quadratic_nw}
		\begin{split}
			E^q_i & = \begin{cases}	\frac{1}{2} || c_p (d^q_i - y^q_i)||_2^2 ~,& \text{if}~y^q_i < d^q_i, \\
				\frac{1}{2} ||c_h (y^q_i - d^q_i)||_2^2  ~,& \text{if}~d^q_i \le y^q_i. \end{cases}
		\end{split}
	\end{equation} 
	The two propositions that follow provide the gradients of the loss functions with respect to the weights of the network. In both propositions, $i$ is one of the samples, $w_{jk}$ represents a weight in the network between two arbitrary nodes $j$ and $k$ in layers $l$ and $l+1$, 
		\begin{equation}
			a^l_j = g^l_j(z_j^l) = \frac{\partial (z^l_k) }{\partial w_{jk}}
		\label{eq:alj}
		\end{equation}
is the activation function value of node $j$, and 
	\begin{equation}
	\label{eq:delta}
	\begin{split}
		\delta_j^l & = \frac{\partial E_i^q }{\partial z_j^l} \\
		& = \frac{\partial E_i^q }{\partial a_j^l} \frac{\partial a_j^l }{\partial z_j^l } \\
		& = \frac{\partial E_i^q }{\partial a_j^l} ({g}^l_j)' (z_j^l).
	\end{split}
	\end{equation}
Also, let
	\begin{equation}
	\label{eq:deltaph}
	\begin{split}
		\delta_j^l(p) & = c_p (g^l_j)'(z^l_j) \\
		\delta_j^l(h) & = c_h (g^l_j)'(z^l_j), 
	\end{split}
	\end{equation}
	denote the corresponding $\delta_j^l$ for the shortage and excess cases, respectively. Proofs of both propositions are provided in Appendix \ref{sec:proofs}.	

	\begin{prop}
	The gradient with respect to the weights of the network for loss function \eqref{original_nw} is:
	\begin{equation} 
		\label{eq:L1_gradient}
		\frac{\partial E^q_i}{\partial w_{jk}} = \begin{cases}
			a^l_j \delta_j^l(p) & \text{if}~y^q_i < d^q_i, \\ 
			a^l_j \delta_j^l(h) & \text{if}~d^q_i \le y^q_i.
		\end{cases}
	\end{equation} 
	\label{prop:gradient1}
	\end{prop}
	\begin{prop}
	The gradient with respect to the weights of the network for loss function \eqref{quadratic_nw} is:
	\begin{equation} 
	\label{eq:L2_gradient}
	\begin{split}
	\frac{\partial E^q_i}{\partial w_{jk}} &=  \left\{\begin{matrix}
	( d^q_i - y^q_i) a^l_j \delta_j^l(p),  & \text{if}~y^q_i < d^q_i\\ 
	( y^q_i - d^q_i) a^l_j \delta_j^l(h),  & \text{if}~d^q_i \le y^q_i.
	\end{matrix}\right. \\
	\end{split}
	\end{equation} 	
	\label{prop:gradient2}
	\end{prop}
	Our deep learning algorithm uses gradient \eqref{eq:L1_gradient} and sub-gradient \eqref{eq:L2_gradient} under the proposed loss functions \eqref{original_nw} and  \eqref{quadratic_nw}, respectively, to iteratively update the weights of the networks.
	In order to obtain the new weights, an SGD algorithm with momentum is called, with a fixed momentum of 0.9.
	This gives us two different DNN models, using the linear loss function \eqref{original_nw} and the quadratic loss function \eqref{quadratic_nw}, which we call  DNN-$\ell_1$ and DNN-$\ell_2$, respectively.

	In order to obtain a good structure for the DNN network, we follow HyperBand algorithm \cite{li2016hyperband}. In particular, we generate 100 fully connected networks with random structures. In each, the number of hidden layers is randomly selected as either two or three (with equal probability). Let $nn_l$ denote the number of nodes in layer $l$; then $nn_1$ is equal to the number of features. The number of nodes in each hidden layer is selected randomly based on the number of nodes in the previous layer. For networks with two hidden layers, we choose $nn_2 \in [0.5nn_1,3nn_1]$, $nn_3 \in [0.5nn_2,nn_2]$, and $nn_4 = 1$. Similarly, for networks with three hidden layers, $nn_2 \in [0.5nn_1,3nn_1]$, $nn_3 \in [0.5nn_2,2nn_2]$, $nn_4 \in [0.5nn_3,nn_3]$, and $nn_5 = 1$. 
	The $nn_l$ values are drawn uniformly from the ranges given.
	For each network, the learning rate and regularization parameters are drawn uniformly from $[10^{-2}, 10^{-5}]$. In order to select the best network among these, following the HyperBand algorithm, we train each of the 100 networks for one epoch (which is a full pass over the training dataset), obtain the results on the test set, and then remove the worst $10\%$ of the networks. We then run another epoch on the remaining networks and remove the worst $10\%$. This procedure iteratively repeats to obtain the final best networks.

	\section{Numerical Experiments}\label{sec:Numerical Experiments}
	In this section, we discuss the results of our numerical experiments. In addition to implementing our deep learning models (DNN-$\ell_1$ and DNN-$\ell_2$), we implemented the EQ model by  \cite{Thiele}, modifying it so that first the demand observations are clustered according to the features and then EQ is applied to each cluster.
	We also implemented the LML and KR models by \cite{vahn_News} and the KNN and RF models by \cite{bertsimas2014predictive}, as well as the SEO approach by calculating the classical solution from \eqref{NW_optimal_solution} with parameters $\mu$ and $\sigma$ set to the mean and standard deviation of the training data in each data cluster. 
	We do not include results for EAS since it is dominated by SEO: SEO uses the newsvendor solution based on estimates of $\mu$ and $\sigma$, whereas EAS simply sets the solution equal to the estimate of $\mu$.
	In order to compare the results of the various methods, the order quantities were obtained with each algorithm and the corresponding cost function 
	$$ \text{cost} = \sum_{i=1}^{n} \sum_{q=1}^{m} \left[c_p (d_{i}^q - y^q_{i})^+ + c_h (y^q_{i} - d^q_{i})^+ \right] $$
	was calculated. 
	
	All of the deep learning experiments were done with TensorFlow 1.4.0  (\cite{abadi2016tensorflow}) in Python. 
	Note that the deep learning, LML, KR, KNN, and RF algorithms are scale dependent, meaning that the tuned parameters of the problem for a given set of cost coefficients do not necessarily work for other values of the coefficients. 
	Thus, we performed a separate tuning for each set of cost coefficients.
	In addition, we translated the categorical data features to their corresponding binary representations (using one-hot encoding). These two implementation details improve the accuracy of the learning algorithms. 
	All computations were done on 16 cores machines with cores of 1.8 GHz computation power and 32 GB of memory.
		
	In what follows, we demonstrate the results of the seven algorithms in three separate experiments. 
	First, in Section \ref{sec:toy_example_section}, we conduct experiments on a very small data set in order to illustrate the differences among the methods.
	Second, the results of the seven algorithms on a real-world dataset are presented in Section \ref{sec:Numerical_Experiments_Real}. 
	Finally, in Section \ref{sec:Numerical_experiments_simulation}, to determine the conditions under which deep learning outperforms the other algorithms on larger instances, we present the results of the seven approaches on several randomly generated datasets.
	
	\linespread{1}
	
	\subsection{Small Data Set}\label{sec:toy_example_section}
	Consider the small, single-item instance whose demands are contained in Table \ref{toy_problem}.
	\begin{table}[]
		\centering
		\caption{Demand of one item over three weeks.}
		\small %
		\label{toy_problem}
		\begin{tabular}{llllllll}
			& Mon & Tue & Wed & Thu & Fri & Sat & Sun \\ \cline{2-8}
			\multicolumn{1}{l|}{Week 1} & 1   & 2   & 3   & 4   & 3   & 2   & 1   \\
			\multicolumn{1}{l|}{Week 2} & 6   & 10  & 12  & 14  & 12  & 10  & 10  \\
			\multicolumn{1}{l|}{Week 3} & 3   & 6   & 8   & 9   & 8   & 6   & 5  \\ 
		\end{tabular}
	\end{table}

	\linespread{2}

	In order to obtain the results of each algorithm, the first two weeks are used for training data and the third week is used for testing. 
	To train the corresponding deep network, a fully connected network with one hidden layer is used. 
	The network has eight binary input nodes for the day of week and item number. The hidden layer contains one sigmoid node, and in the output layer there is one inner product function. 
	Thus, the network has nine variables. 
	
	Table \ref{toy_result} shows the results of the seven algorithms. The first column gives the cost coefficients. Note that we assume $c_p \ge c_h$ since this is nearly always true for real applications. The table first lists the actual demand for each day, repeated from Table \ref{toy_problem} for convenience. For each instance (i.e., set of cost coefficients), the table lists the order quantity generated by each algorithm for each day. The last column lists the total cost of the solution returned by each algorithm, and the minimum costs for each instance are given in bold. 
{   
\linespread{1}	
	\begin{table}[]
		\centering
		\caption{Order quantity proposed by each algorithm for each day and the corresponding cost.
			The bold costs indicate the best newsvendor cost for each instance.
		}
		\label{toy_result}
		\small 
		\begin{tabular}{ll|ccccccc|c}
			\hline 
			& & \multicolumn{7}{c|}{Day \& Demand} \\
			($c_p,c_h$)& Algorithm   & Mon & Tue & Wed & Thu & Fri & Sat & Sun &    Cost  \\ \hline
			& True demand & 3 & 6 & 8 & 9 & 8 & 6 & 5 \\ \hline
			\multirow{5}{*}{(1,1)} 
			& DNN-$\ell_2$ & 3.5 & 6.0  & 7.5  & 9.0  & 7.5  & 6.5  & 5.5  & {\bf  2.5}  \\	
			& DNN-$\ell_1$& 	4.6& 	6.0	& 8.5& 	9.0& 	8.6	& 5.6& 	5.6	& 2.9 \\				
			& EQ & 1.0 & 2.0  & 3.0  & 4.0  & 3.0  & 2.0  & 1.0  & 29.0 \\
			& LML & 1.3 & 2.2  & 3.1  & 4.0  & 4.9  & 5.8  & 6.7  & 20.3 \\
			& SEO  & 3.5 & 6.0  & 7.5  & 9.0  & 7.5  & 6.5  & 5.5  & {\bf 2.5}  \\		
			& KR  & 4.0 & 6.0  & 6.0  & 6.0  & 6.0  & 4.0  & 4.0  & 11.0  \\			
			& KNN  & 6.0 & 6.0  & 6.0  & 6.0  & 6.0  & 6.0  & 6.0  & 11.0  \\					 
			& RF  & 6.0 & 6.0  & 6.0  & 6.0  & 6.0  & 6.0  & 6.0  & 11.0  \\				
			\hline 
			\multirow{5}{*}{(2,1)} 
			& DNN-$\ell_2$ & 5.8 & 7.3  & 8.9  & 9.7  & 8.9  & 8.1  & 7.0  & 10.7 \\
			& DNN-$\ell_1$ &	6.0&	6.0&	7.9	 & 9.3&	9.3&	6.4&	6.1&{\bf 	5.9}
			 \\							
			& EQ & 6.0 & 10.0 & 12.0 & 14.0 & 12.0 & 11.0 & 10.0 & 30.0 \\
			& LML & 6.0 & 7.0  & 8.0  & 9.0  & 10.0 & 11.0 & 12.0 & 18.0 \\ 
			& SEO  & 5.0 & 8.4  & 10.2 & 12.0 & 10.2 & 9.2  & 8.2  & 18.5 \\		
			& KR  & 6.0 & 10.0  & 12.0  & 14.0  & 12.0  & 11.0  & 10.0  & 30.0  \\	
			& KNN  & 10.0 & 6.0  & 10.0  & 10.0  & 10.0  & 10.0  & 10.0  & 25.0  \\	
			& RF  & 6.0 & 10.0  & 10.0  & 10.0  & 11.0  & 11.0  & 11.0  & 24.0  \\								
			\hline
			\multirow{5}{*}{(10,1)}
			& DNN-$\ell_2$ & 5.6 & 9.3  & 11.2 & 13.1 & 11.2 & 10.2 & 9.2  & 24.6 \\
			& DNN-$\ell_1$&	6.9&	10.2&	10.2&	10.2&	10.2&	10.2&	10.2&	{\bf 22.8} \\			
			& EQ & 6.0 & 10.0 & 12.0 & 14.0 & 12.0 & 11.0 & 10.0 & 30.0 \\
			& LML & 8.0 & 10.0 & 12.0 & 14.0 & 16.0 & 18.0 & 20.0 & 53.0 \\ 
			& SEO  & 8.2 & 13.6 & 16.0 & 18.4 & 16.0 & 15.0 & 14.0 & 56.2 \\	
			& KR  & 6.0 & 10.0  & 12.0  & 14.0  & 12.0  & 11.0  & 10.0  & 30.0  \\
			& KNN  & 6.0 & 10.0  & 12.0  & 12.0  & 12.0  & 11.0  & 10.0 & 39.0 \\
			& RF  & 12.0 & 12.0  & 12.0  & 14.0  & 12.0  & 12.0  & 12.0  & 41.0  \\			
			\hline
			\multirow{5}{*}{(20,1)} 
			& DNN-$\ell_2$ & 5.8 & 9.6  & 11.6 & 13.5 & 11.6 & 10.6 & 9.6  & {\bf 27.2} \\		
			& DNN-$\ell_1$&	6.1&	11.3&	11.3&	11.3&	11.3&	11.3&	11.3&	28.6 \\
			& EQ & 6.0 & 10.0 & 12.0 & 14.0 & 12.0 & 11.0 & 10.0 & 30.0 \\
			& LML & 8.0 & 10.0 & 12.0 & 14.0 & 16.0 & 18.0 & 20.0 & 38.0	\\
			& SEO  & 9.4 & 15.4 & 18.1 & 20.8 & 18.1 & 17.1 & 16.1 & 70.1\\ 
			& KR  & 10.0 & 12.0  & 14.0  & 14.0  & 14.0  & 12.0  & 11.0  & 42.0  \\ 
			& KNN  & 14.0 & 14.0  & 14.0  & 14.0  & 14.0  & 14.0  & 14.0  & 53.0 \\		
			& RF  & 14.0 & 14.0  & 14.0  & 14.0  & 14.0  & 14.0  & 14.0  & 53.0 \\			
			\hline
		\end{tabular}
	\end{table}
	}
	
	First consider the results of the EQ algorithm. 
	The EQ algorithm uses $c_h$ and $c_p$ and returns the historical data value that is closest to the $\alpha$th fractile, where $\alpha = c_p/(c_h+c_p)$. 
	In this data set, there are only two observed historical data points for each day of the week.
	In particular, for $c_p/c_h \le 1$, the EQ algorithm chose the smaller of the two demand values as the order quantity, and for $c_p/c_h > 1$, it chose the larger value. 
	Since the testing data vector is nearly equal to the average of the two training data vectors, the difference between EQ's output and the real demand values is quite large, and consequently so is the cost. 
	This is an example of how the EQ algorithm can fail when the historical data are volatile.

	Consider the KNN algorithm. Since there are only two weeks of historical data, we opt to use all possible historical records without any validation and set $k=14$. KNN gets the $k$ historical records that are nearest to the new observation, each with a weight of $\frac1k$, and then chooses the point that weighted SAA selects. The demand of that point is the order quantity. 
	So, as $c_p/c_h$ increases, it selects larger values. 
	However, the demands during the third week (the testing set) are close to the mean demand of the first two weeks (the training set); therefore, the increased order quantity chosen by KNN turns out to be too large.
	Similarly, in RF we select 2000 forests, and in KR we select $h=0.5$ and use all data from the two weeks of the training set. Since both algorithms work with sorted demands, once $c_p/c_h$ increases, they select larger demands from the training sets. 
	Therefore, RF and KR also results in large cost values, for similar reasons as KNN.

	Now consider the results of the SEO algorithm. For the case in which  $c_h = c_p$ (which is not particularly realistic), SEO's output is approximately equal to the mean demand, which happens to be close to the week-3 demand values. This gives SEO a cost of 2.5, which ties DNN-$\ell_2$ for first place. For all other instances, however, the increased value of $c_p/c_h$ results in an inflated order quantity and hence a larger cost.
	%
	
	Finally, both DNN-$\ell_1$ and DNN-$\ell_2$ outperform the LML algorithm by \cite{vahn_News}, because LML uses a linear kernel, while DNN uses both a linear and non-linear kernel. 
	Also, there are only two features in this data set, so LML has some difficulty to learn the relationship between the inputs and output. 
	Finally, the small quantity of historical data negatively affects the performance of LML.
	
	This small example shows some conditions under which DNN outperforms the other three algorithms.
	In the next section we show that similar results hold even for a real-world dataset.
	
	\subsection{Real-World Dataset}\label{sec:Numerical_Experiments_Real}
	We tested the seven algorithms on a real-world dataset consisting of basket data from a retailer in 1997 and 1998 from \cite{FoodMart}. 
	There are 13170 records for the demand of 24 different departments in each day and month, of which we use $75\%$ for training and validation and the remainder for testing. 
	The categorical data were transformed into their binary equivalents, resulting in 43 input features. 
	
\begin{figure}[]
		\centerline{\includegraphics[width=12cm]{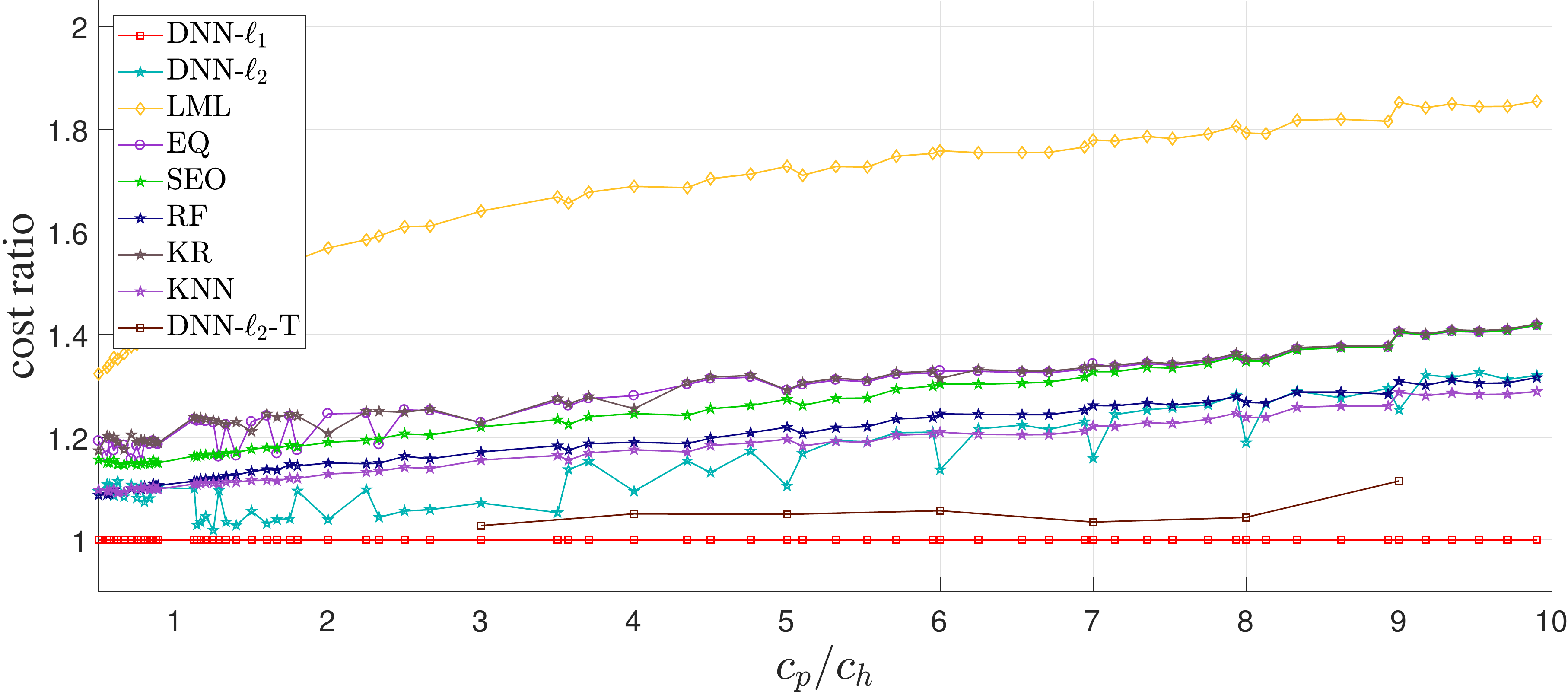}	}
		\caption{Ratio of each algorithm's cost to DNN-$\ell_1$ cost on a real-world dataset.}
		\label{fig:real_data_resultx}	
	\end{figure}

	The results of each algorithm for 100 values of $c_p$ and $c_h$ are shown in Figure \ref{fig:real_data_resultx}. 
	In the figure, the vertical axis shows the normalized costs, i.e., the cost value of each algorithm divided by the corresponding DNN-$\ell_1$ cost.
	The horizontal axis shows the ratio $c_p/c_h$ for each instance. As before, most instances use $c_p \ge c_h$ to reflect real-world settings, though a handful of instances use $c_p < c_h$ to test this situation as well. 
	
	As shown in Figure \ref{fig:real_data_resultx}, for this data set, the DNN-$\ell_1$ and DNN-$\ell_2$ algorithms both outperform the other three algorithms for every value of $c_p / c_h$. 
	Among the three remaining algorithms, the results of the KNN and RF algorithms are the closest to those of DNN. 
	On average, their corresponding cost ratios are 1.15 and 1.16, whereas the ratios for EQ, LML, KR, and SEO are 1.26, 1.53, 1.16, 1.26, and 1.23, respectively. 
	The average cost ratio of DNN-$\ell_2$ is 1.13.
	However, none of the other approaches are stable; their cost ratios increase with the ratio $c_p/c_h$.
	
	DNN-$\ell_2$ requires more tuning than DNN-$\ell_1$, but the DNN-$\ell_2$ curve in Figure~\ref{fig:real_data_resultx} does not reflect this additional tuning. 
	The need for additional tuning is suggested by the fact that DNN-$\ell_2$'s loss value increases as $c_p$ or $c_h$ increase, suggesting that it might need a smaller learning rate (to avoid big jumps) and a larger regularization coefficient $\lambda$ (to strike the right balance between cost and over-fitting).\ Thus, tuning DNN-$\ell_2$ properly would require a larger search space of the learning rate and $\lambda$, which would make the procedure harder and more time consuming. In our experiment, we did not expend this extra effort; instead, we used the same procedure and search space to tune the network for both DNN-$\ell_1$ and DNN-$\ell_2$, in order to compare them fairly. 
	
	Nevertheless, it is worth investigating how the performance of DNN-$\ell_2$ could be improved if it is tuned more thoroughly. To that end, we selected integer values of $c_p/c_h = 3,\ldots,9$, and for each value, we applied more computational power and tuned the parameters using a grid search. We fixed the network as $[43,350,100,1]$, tested it with 702 different parameters, and selected the best test result among them. The grid search procedure is explained in detail in Appendix \ref{appendix basket_grid_search}. The corresponding result is labeled as DNN-$\ell_2$-T in Figure \ref{fig:real_data_resultx}. As the figure shows, this approach has better results than the original version of DNN-$\ell_2$; however, DNN-$\ell_1$ is still better.

	The DNN algorithms execute more slowly than some of the other algorithms. 
	For the basket dataset, the SEO and EQ algorithms each execute in about 10 seconds. 
	The DNN algorithm requires about 50 seconds (on a relatively large network, e.g., $[43, 90, 150, 56, 1]$) for each epoch of training, while the LML, KR, KNN, and RF algorithms require on average, respectively, about 40 seconds (per regularization value), 15 seconds (per bandwidth), 5 seconds (for a given $k$), and 4 seconds (per tree) for training for a given $c_p$ and $c_h$. As the size of the search space for hyper-parameter tuning increases, so does the training time  for DNN, LML, KR, RF, and KNN. 
	For LML, we tested 30 different bandwidths---$2^h, h \in \{-20, \dots, 10\}$---which resulted in 1200 seconds of training, on average.
	For KR, we tested bandwidth values of $10^{-5}, 10^{-4}, 10^{-3}, 10^{-2}, 0.05, 0.1,$ and $0.25$, with a total time of 110 seconds on average. KNN needs to tune $k$, for which we tested six values---$5, 10, 15, 50, 100,$ and $200$---which took 30 seconds on average. Similarly, for RF we tested five forest sizes---$10, 20, 50, 100,$ and $150$---which resulted in 1320 seconds of training on average.
	For DNN, we used the HyperBand algorithm to tune the network. We tested several different values of each of the hyper-parameters (as explained at the end of Section \ref{sec:methods}), 
resulting in a total of 881 epochs, which took 12.25 hours of training on average.  
The best network runs for 16 epochs, which took 600 seconds on average. 
Table~\ref{table:tune} summarizes the hyper-parameters used during the tuning process for each method, and their approximate computation times. Note that the times reported in the table are for one instance of the basket dataset, i.e., one value of $c_p/c_h$.
	
	\linespread{1}

	\begin{table}
	\caption{Summary of hyper-parameter (HP) tuning process for each method. Times reported are approximate training times for a single problem instance.}
	\label{table:tune}
		\centering
		\begin{tabular}{l|ccccc}
			& & & & Approx.~Avg. & Approx. \\
			& & \# HP Values & & Training Time & Total Training \\
			& \# HP & Tested & HP Values Tested & per HP (sec) & Time (sec) \\ \hline
			SEO & --   &   -- &  &     &  10   \\ 
			EQ & --  &   -- &  &     &  10    \\ 
			LML & 1  & 30 & $2^h, h \in \{-20, \dots, 10\}$ &  40  &  1200   \\ 
			KR & 1 & 7 & $\{10^{-5}, 10^{-4}, 10^{-3}, 10^{-2}, 0.05, 0.1, 0.25\}$ &  15  &  105  \\
			KNN & 1 & 6 & $\{5, 10, 15, 50, 100, 150\}$ &  5  &  30  \\
			RF & 1 & 5 & $\{10, 20, 50, 100, 150\}$ &  4~(per tree)  &  1320  \\
			DNN & 4 & 100 & (see Section~\ref{sec:methods}) &  600  &  44,050  \\ \hline
		\end{tabular}		
	\end{table}

	\linespread{2}

	On the other hand, DNN and LML algorithms execute in less than one second, i.e., once the network is trained, the methods generate order quantities for new instances very quickly. In contrast, KR, KNN, and RF required approximately 15, 5, and $4t$ seconds, respectively, for inference, where $t$ is the number of trees that is selected.
	
	Since tuning the DNN hyper-parameters can be time-consuming, in Appendix \ref{sec:appnd:tune-free-structure} we propose a simple tuning-free network for the newsvendor problem.

\begin{figure}[]
	\centering	
	\begin{subfigure}[b]{0.3\textwidth}		
		\centering		
		\includegraphics[width=4.4cm]{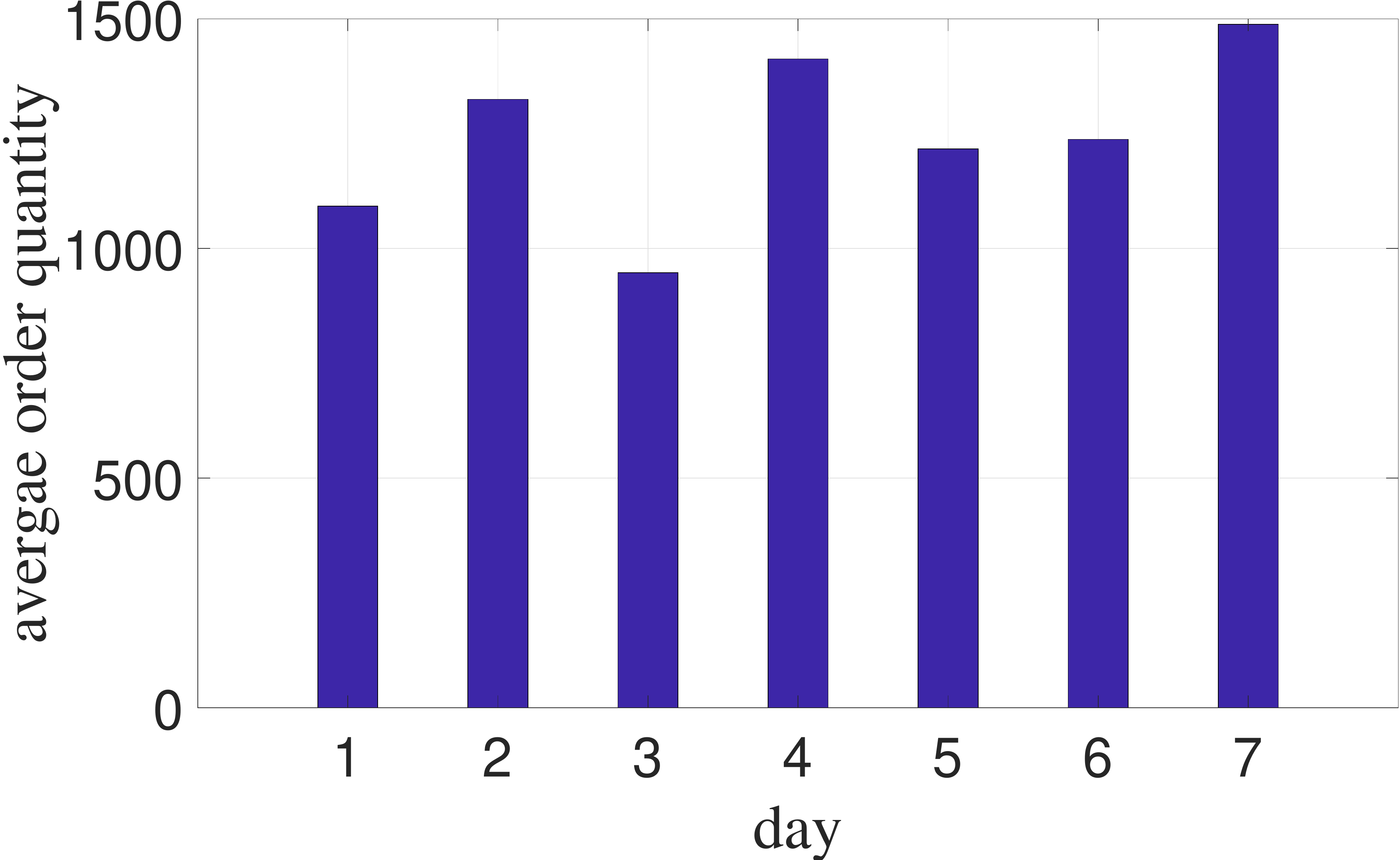}	
		\caption{}
		\label{fig:featue_day}			
	\end{subfigure}	
	\begin{subfigure}[b]{0.3\textwidth}		
		\centering			
		\includegraphics[width=4.4cm]{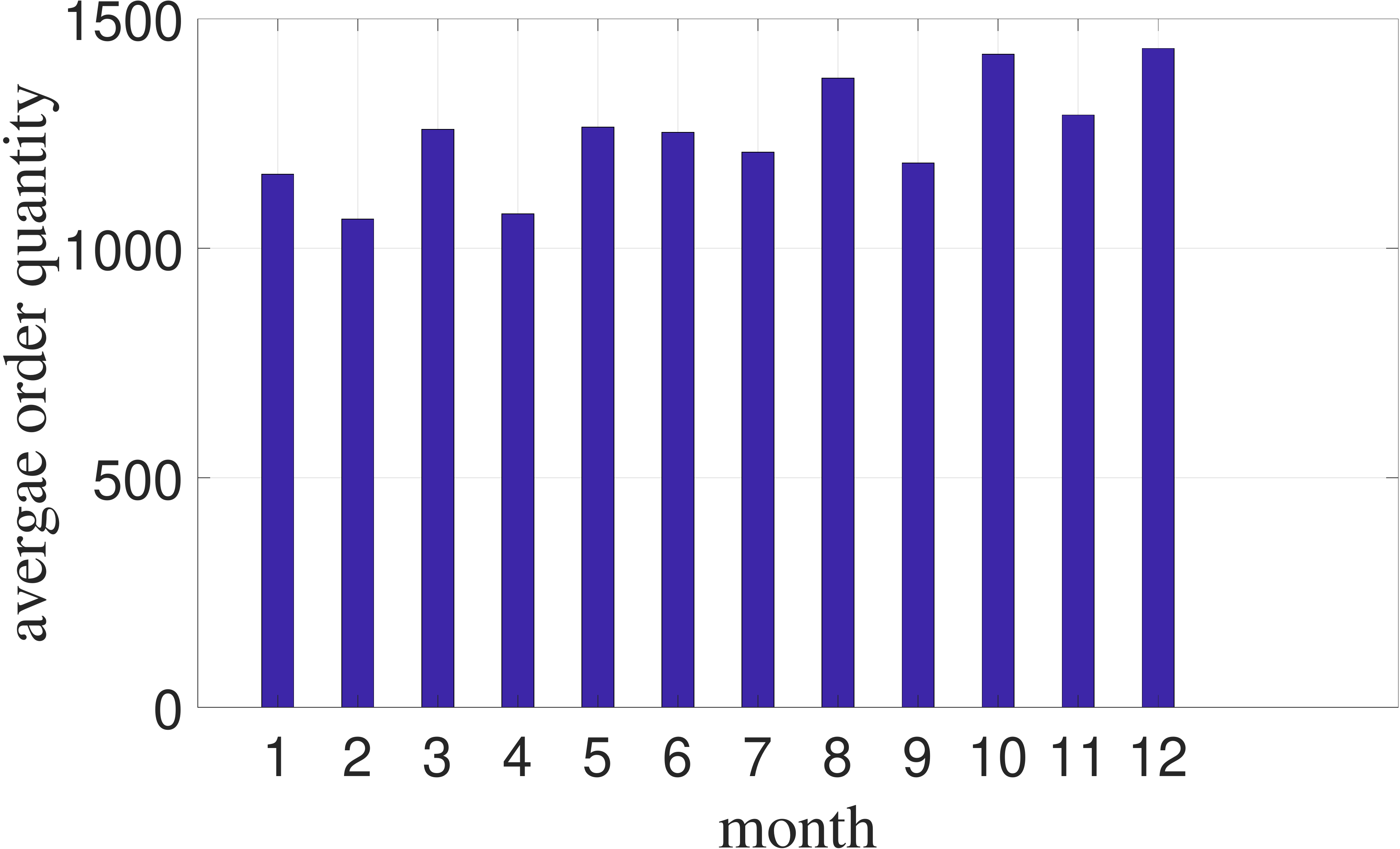}
		\caption{}
		\label{fig:featue_month}			
	\end{subfigure}	
	\begin{subfigure}[b]{0.3\textwidth}		
		\centering		
		\includegraphics[width=4.4cm]{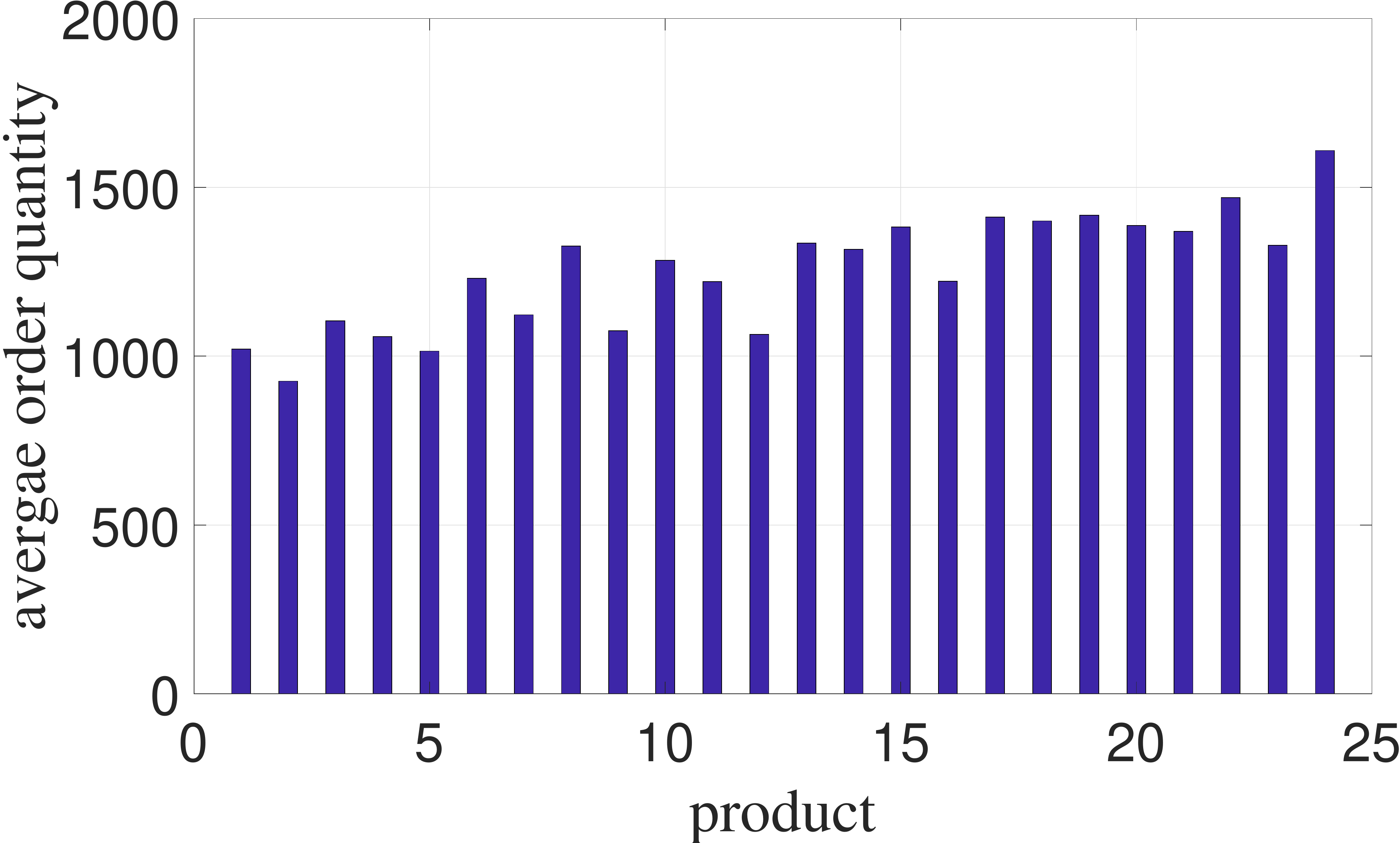}	
		\caption{}
		\label{fig:featue_product}			
	\end{subfigure}				
	\caption{The effect each feature on the order quantity for uniformly distributed data with 100 clusters.}
	\label{fig:featue_effect}	
\end{figure}

Finally, we performed a small experiment to provide some intuition about which features have the most impact on the order quantity. In particular, we calculated the order quantity for each of the $7\times 12\times 24 = 2016$ possible combinations of the feature values, using the DNN model tuned for a uniform distribution with 100 clusters. For each individual feature value, we calculated the average order quantity; these are plotted in Figure~\ref{fig:featue_effect}. From the figure it is evident that---for this data set---the order quantity is affected most strongly by the product category, then by the day of the week, and then by the month of the year. The average order quantity ranges (max $-$ min) for the product, day, and month are 682.9, 540.7, and 371.9, respectively. 

This sort of approach could be used to analyze the results of the DNN algorithm for any set of categorical features. The results could be useful to managers attempting to decide whether to use a feature-based approach---including DNN or the other models discussed here---rather than treating the entire data set as a single cluster. For example, if the a supply chain manager for the supermarket data set did not have access to product labels, a feature-based optimization approach would be less valuable, since the day and month features provide less differentiation in the order quantities; in this case, ignoring the features and treating the entire data set as a single cluster would result in less error than it would if product labels were available. Of course, these insights pertain only to this data set. We are not claiming that product is a stronger differentiator  than month in general, but rather illustrating how the DNN model can be used to generate such insights.


	\subsection{Randomly Generated Data}\label{sec:Numerical_experiments_simulation}
	
	In this section we report on the results of an experiment using randomly generated data. This experiment allows us to test the methods on many more instances; however, the disadvantage is that these data are much cleaner than those typically encountered in real supply chains, i.e., they come from a single probability distribution with no noise. This should be kept in mind when interpreting these results. In short, the results in this section indicate that, when the data are non-noisy, all of the methods perform more or less similarly, with some exceptions.  In all cases, DNN's performance is competitive with, if not better than, the other methods; and since it also performs better on messier data sets (e.g., the real-world data set in Section~\ref{sec:Numerical_Experiments_Real}), we recommend its use in general. We now present a more detailed discussion of this experiment.
	
	We conducted tests using five different probability distributions for the demand (normal, lognormal, exponential, uniform, and beta distributions). For each distribution, we generated 257,500 records.  The parameters for the five demand distributions are given in Table~\ref{table:parameters}; these parameters were selected so as to provide reasonable demand values.  All demand values are rounded to the nearest integer.
	Each group of 257,500 records is divided into training and validation (10,000 records) and testing (99 sets, each 2,500 records) sets. 

	\linespread{1.2}

	\begin{table}
	\centering
	\caption{Demand distribution parameters for randomly generated data.}
	\label{table:parameters}
	\begin{tabular}{l|cccc}
		& \multicolumn{4}{c}{Number of Clusters} \\
	Distribution & 1 & 10 & 100 & 200 \\ 
	\hline
	Normal & $\mathcal{N}(50, 10)$ & $\mathcal{N}(50i, 10i)$ & $\mathcal{N}(50i, 5i)$ & $\mathcal{N}(50i, 5i)$ \\
	Lognormal & $\ln\mathcal{N}(2,0.5)$ & $\ln\mathcal{N}(1+0.1(i+1),$ & $\ln\mathcal{N}(0.05(i+1),$ & $\ln\mathcal{N}(0.02(i+1),$ \\
		& & $0.5+0.1(i+1))$ & $0.01(i+1))$ & $0.005(i+1))$ \\
	Exponential & $\exp(10)$ & $\exp(5+2(i+1))$ & $\exp(5+0.2(i+1)$ & $\exp(5+0.05(i+1)$ \\
	Beta & $20\mathcal{B}(1,1)$ & $100\mathcal{B}(0.6(i+1),$ & $100\mathcal{B}(0.1(i+1),$ & $100\mathcal{B}(0.07(i+1),$ \\
		& & $0.6(i+1))$ &  $0.1(i+1))$ & $ 0.07(i+1))$ \\
	Uniform & $\mathcal{U}(1,21)$ & $\mathcal{U}(5(i+1,$ & $\mathcal{U}((i+1),$ & $\mathcal{U}(0.5(i+1),$ \\ 
		& & $15+5(i+1))$ & $15+(i+1))$ & $15+0.5(i+1))$ \\
	\hline
	\end{tabular}
	\end{table}

	\linespread{2}

	In each of the distributions, the data were categorized into clusters, each representing a given combination of features. Like the real-world dataset, we considered three features: the day of the week, month of the year, and department. We varied the number of clusters (i.e., the number of possible combinations of the values of the features) from 1 to 200 while keeping the total number of records fixed at 257,500; thus, having more clusters is the same as having fewer records per cluster.
	In this experiment, an ``instance'' refers to a given combination of demand distribution (normal, exponential, ...) and number of clusters (1, 10, ...). 

	Each problem was solved for $c_p/c_h=5$ using all seven algorithms (including both loss functions for DNN), without assuming any knowledge of the demand distribution. We conducted additional tests using additional $c_p/c_h$ ratios; the results and conclusions were similar, so they are omitted here in the interest of conciseness.
	
	In part, this experiment is designed to model the situation in which the decision maker does not know the true demand distribution. To that end, our implementation of the SEO algorithm assumes the demands come from a normal distribution (regardless of the true distribution for the dataset being tested), since this distribution is used frequently as the default distribution in practice. The other algorithms (DNN, LML, EQ, KNN, KR, and RF) do not assume any probability distribution. 
	Additionally, since we know the underlying demand distributions, we also calculated and reported the optimal solution in each case. 
	The average times required to tune or execute each of the algorithms, per instance, are similar to those in Table~\ref{table:tune}. 

\begin{figure}[]
		\centering
		\begin{subfigure}[b]{0.33\textwidth}	
		\includegraphics[width=5.5cm]{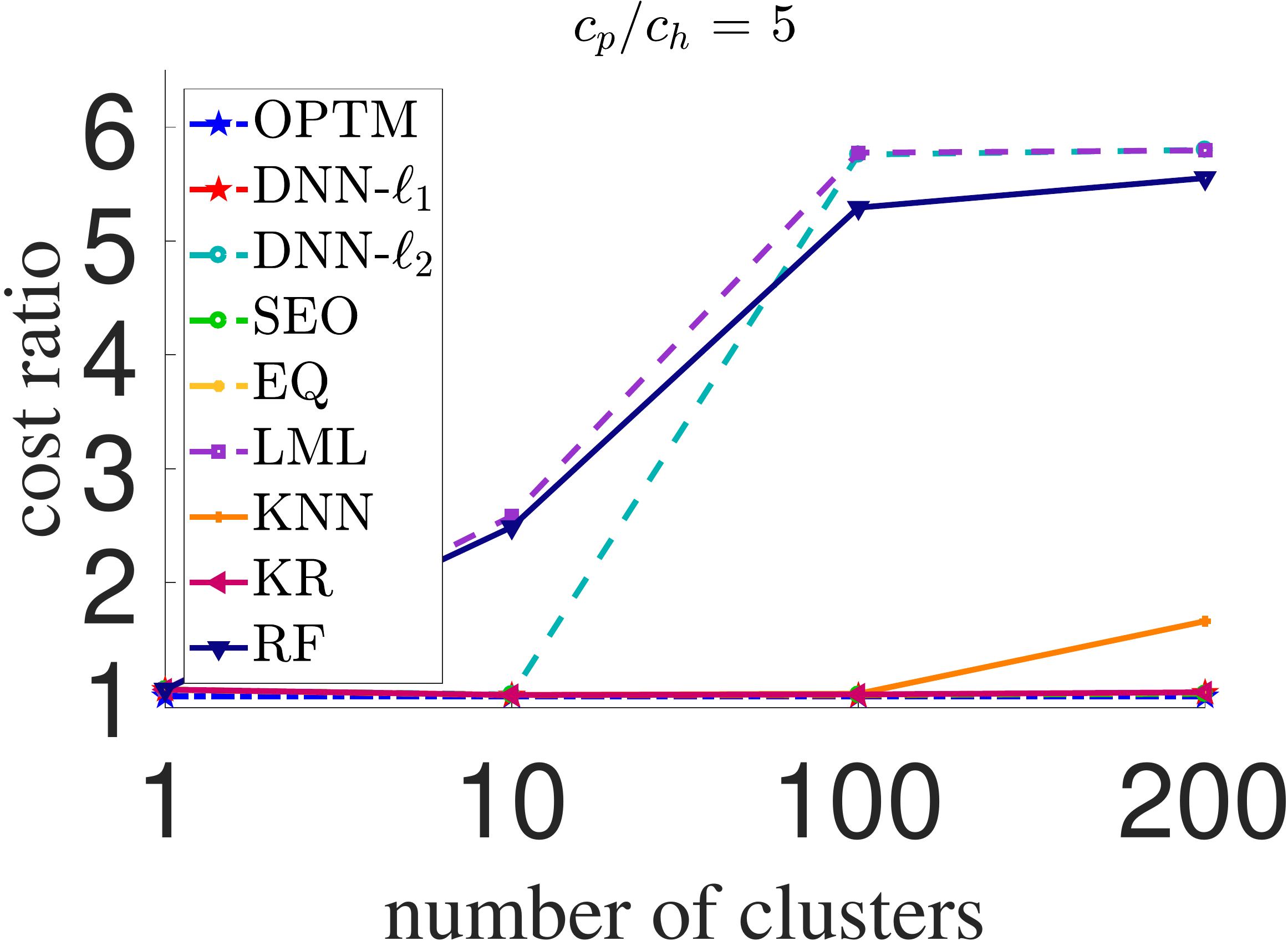}
		\caption{normal}
		\label{fig:simu_data_result_normal}	
		\end{subfigure}	
		\begin{subfigure}[b]{0.33\textwidth}	
		\centering
		\includegraphics[width=5.5cm]{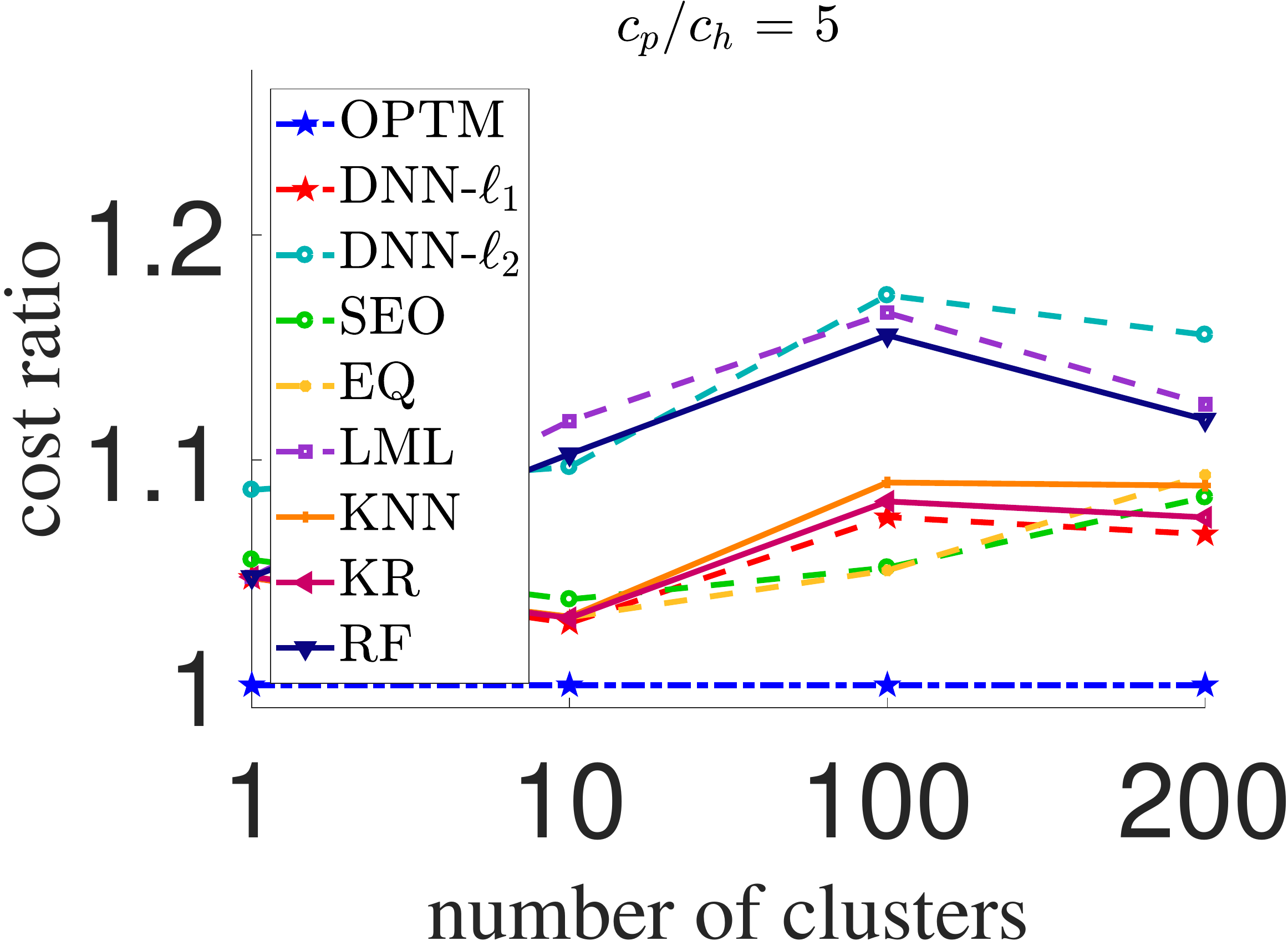}	
		\caption{exponential}
		\label{fig:simu_data_result_exponential}	
		\end{subfigure}	
		\begin{subfigure}[b]{0.33\textwidth}	
		\centering
		\includegraphics[width=5.5cm]{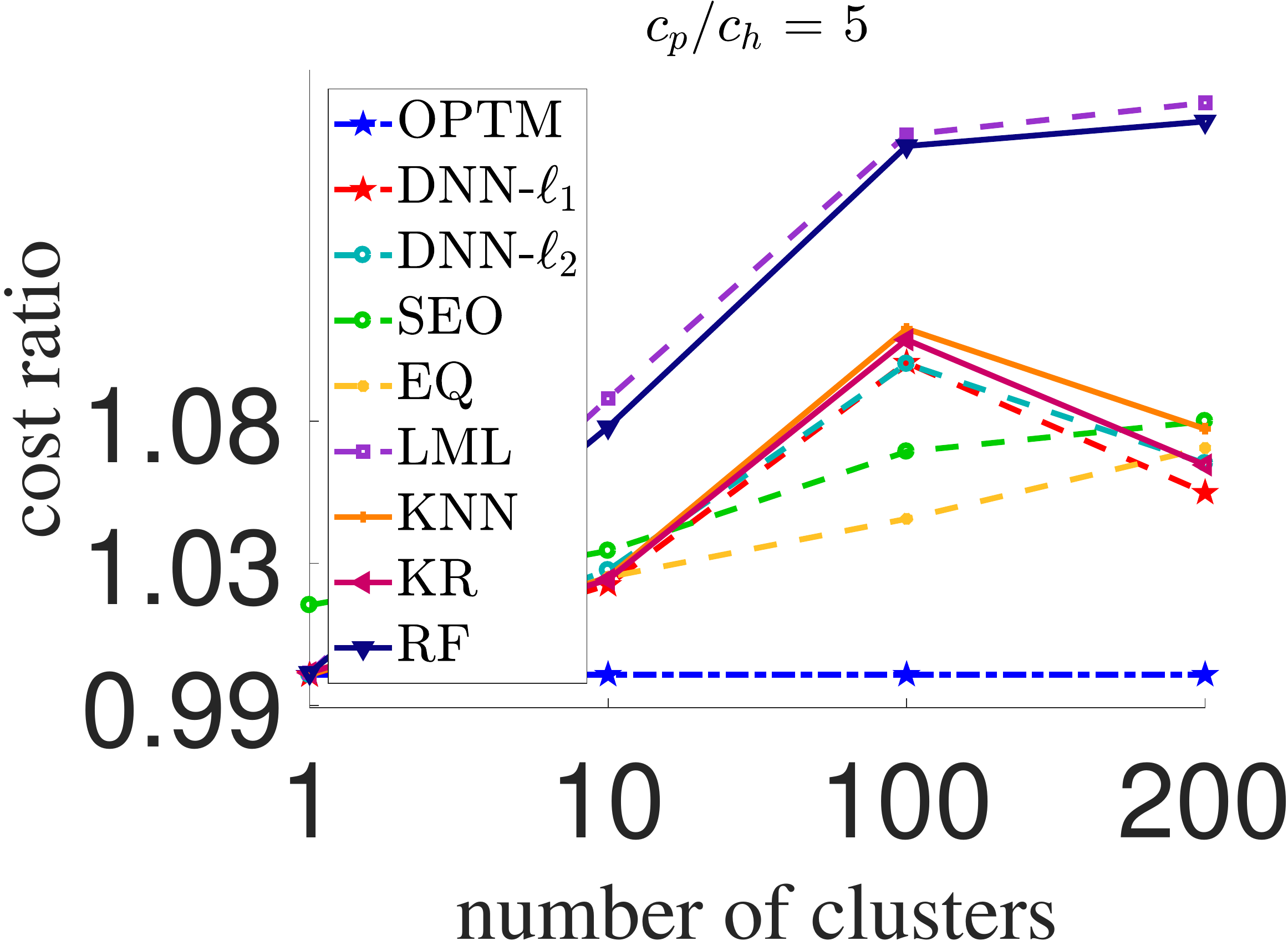}
		\caption{beta}
		\label{fig:simu_data_result_beta}	
		\end{subfigure}	

		\begin{subfigure}[b]{0.33\textwidth}			 	\includegraphics[width=5.5cm]{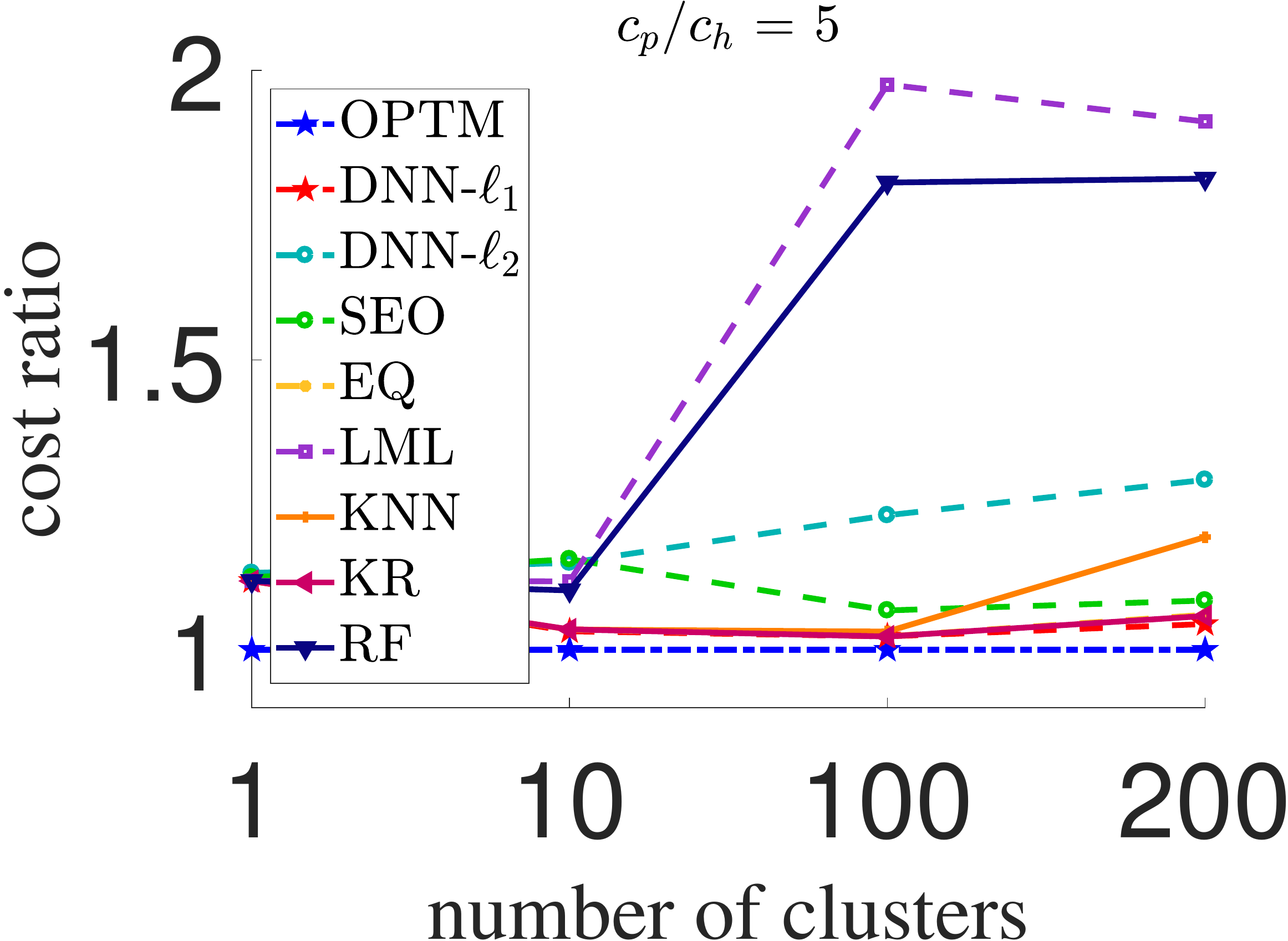}	
		\caption{lognormal}
		\label{fig:simu_data_result_lognormal}	
		\end{subfigure}	
		\begin{subfigure}[b]{0.33\textwidth}	
		\includegraphics[width=5.5cm]{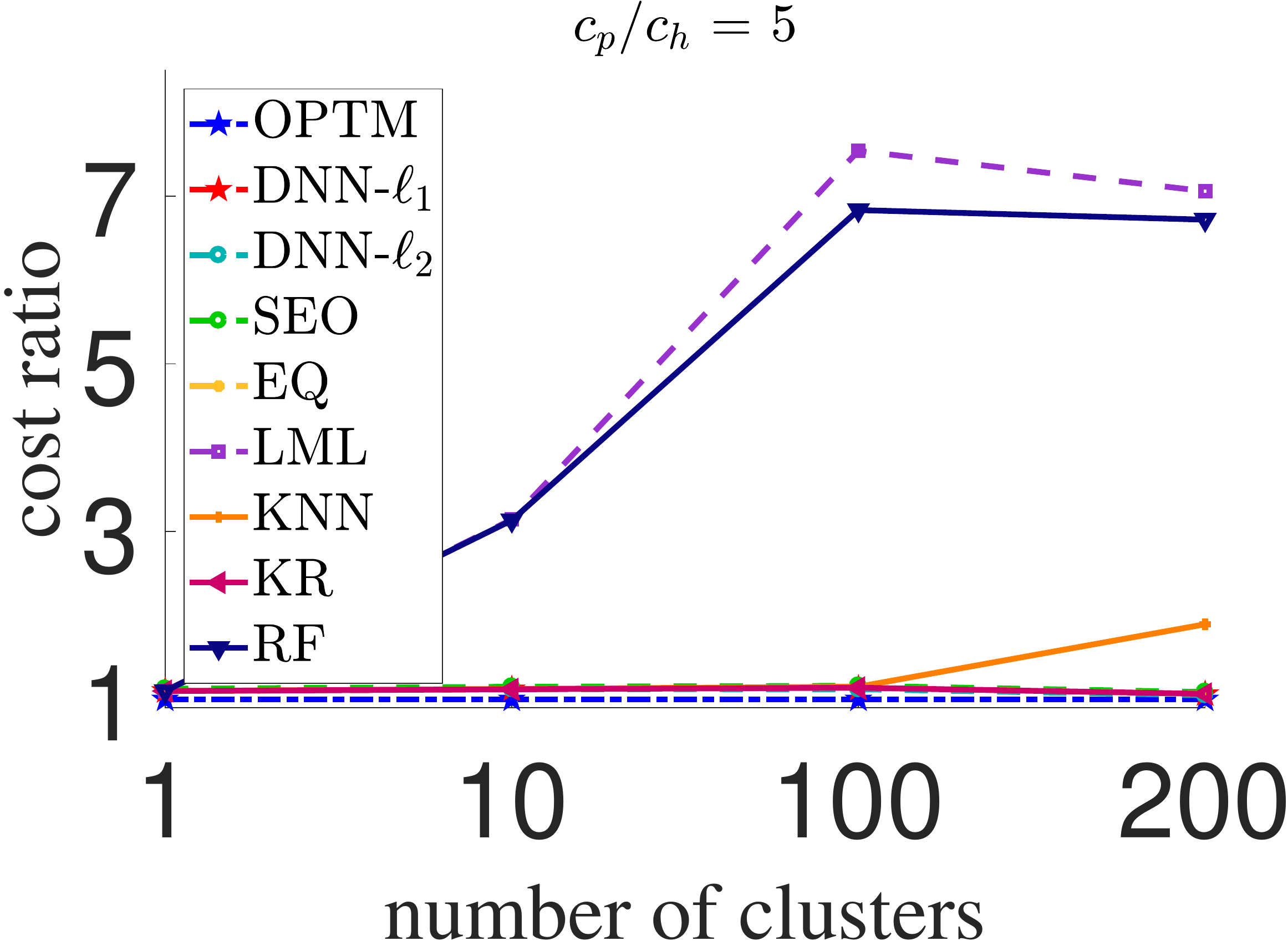}	
		\caption{uniform}
		\label{fig:simu_data_result_uniform}	
		\end{subfigure}
		\caption{Ratio of each algorithm's cost to optimal cost on randomly generated data from each distribution.}					\label{fig:simu_data_result}		
	
	\end{figure}

\begin{figure}[]
		\centering
		\begin{subfigure}[b]{0.32\textwidth}
			\includegraphics[width=5.3cm]{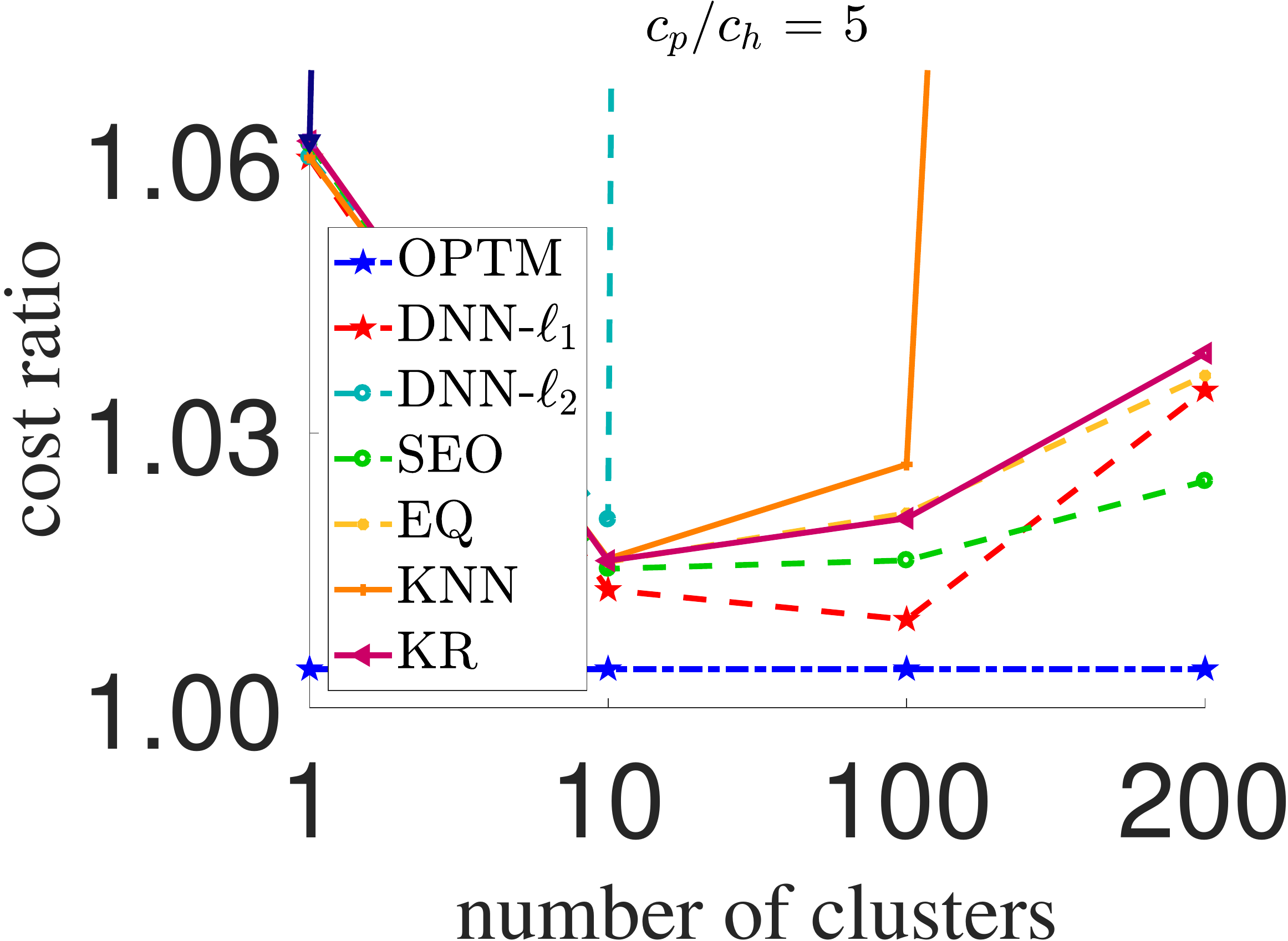}		
			\caption{normal}		
			\label{fig:simu_data_result_normal_magnified}	
		\end{subfigure}	
		\begin{subfigure}[b]{0.32\textwidth}
			\includegraphics[width=5.3cm]{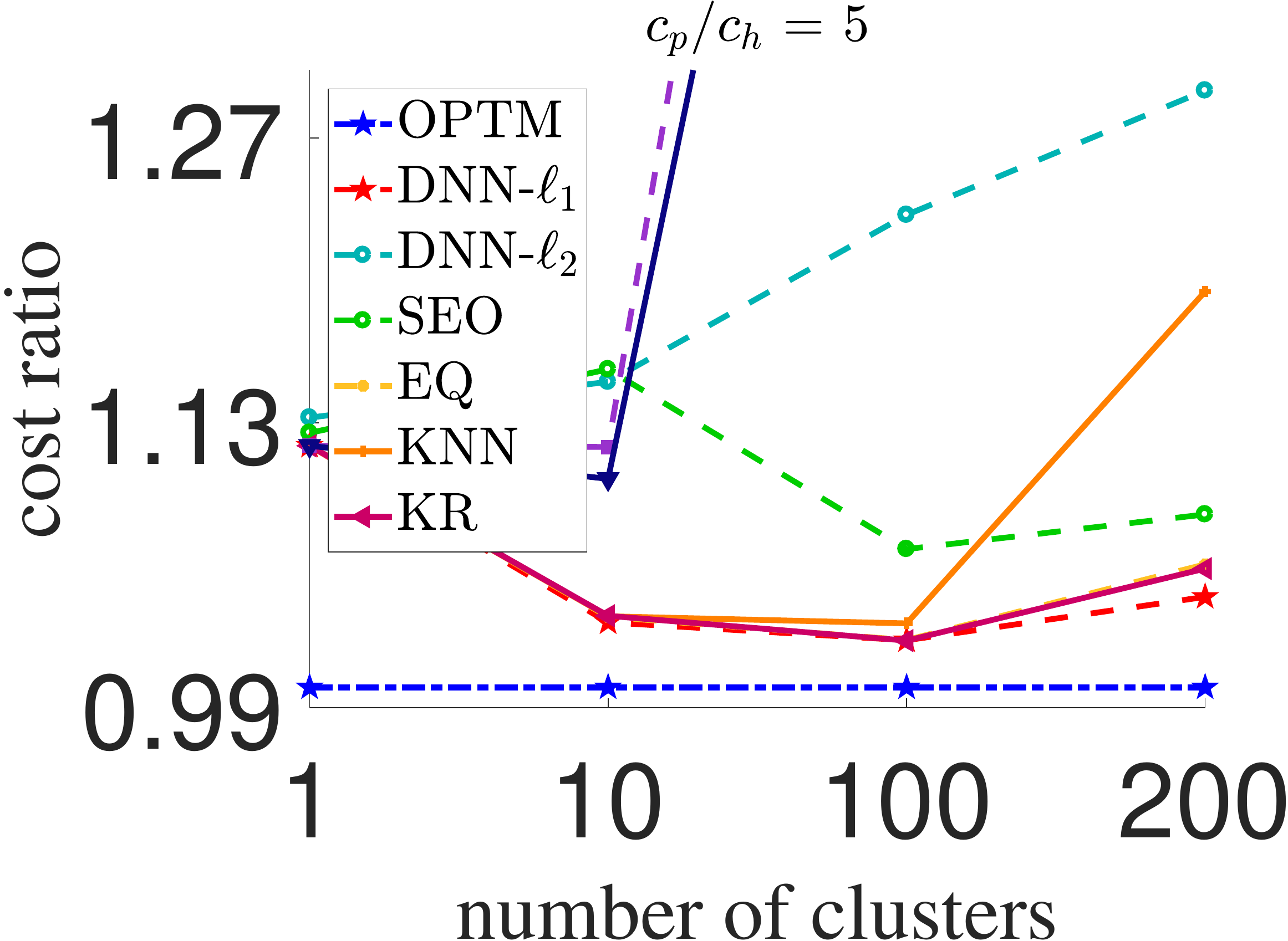}		
			\caption{lognormal}		
			\label{fig:simu_data_result_lognormal_magnified}	
		\end{subfigure}	
		\begin{subfigure}[b]{0.32\textwidth}
			\includegraphics[width=5.3cm]{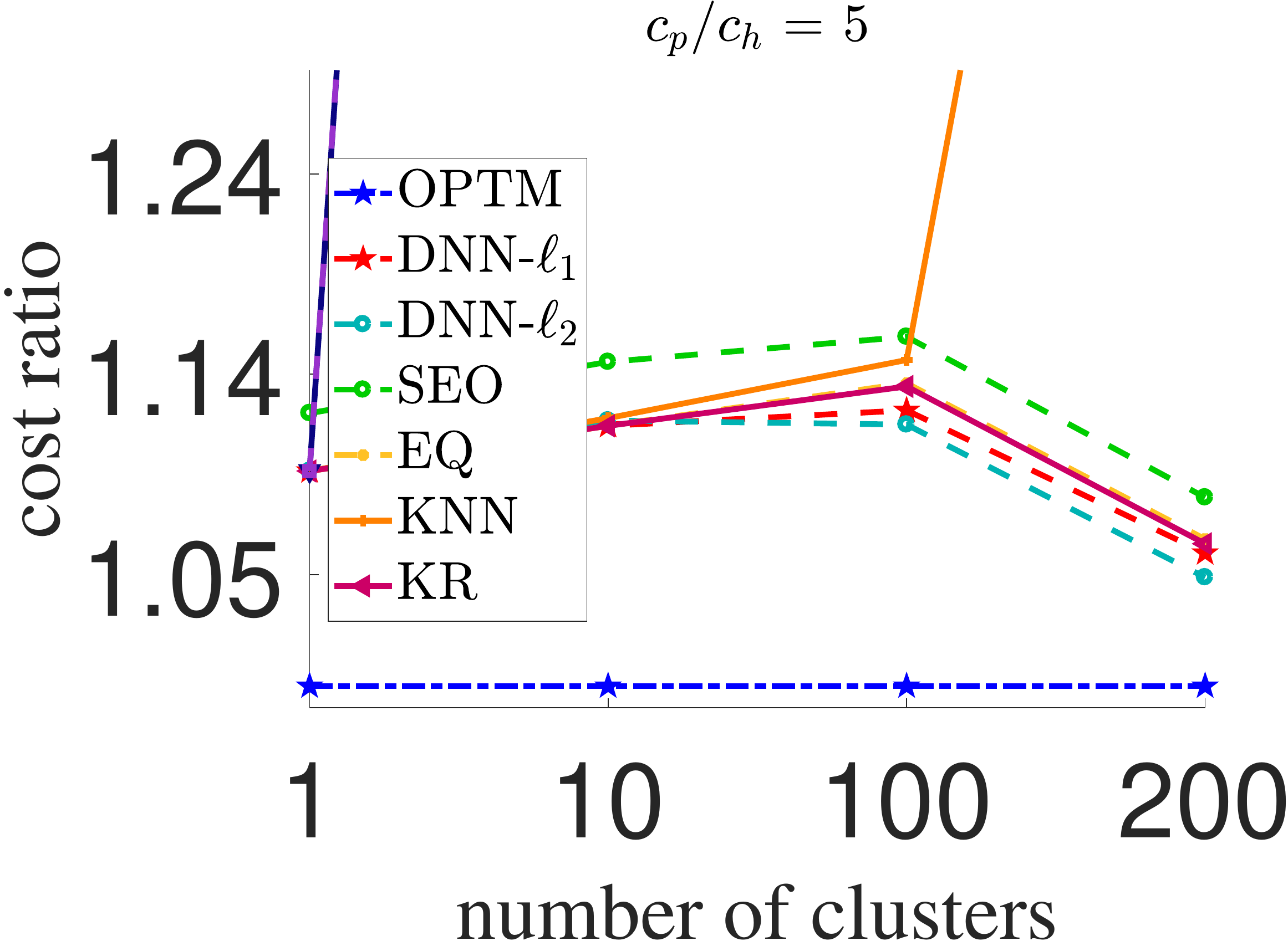}		
			\caption{uniform}								
			\label{fig:simu_data_result_uniform_magnified}	
		\end{subfigure}
		\caption{Magnified results for normal, lognormal, and uniform distributions.}				
		\label{fig:simu_data_result_magnified}	
	\end{figure}

	
	Figure \ref{fig:simu_data_result} plots the average cost ratio (cost divided by optimal cost) for the five distributions. Each point on a given plot represents the average cost (over 99 testing sets) for one instance. Figure \ref{fig:simu_data_result_magnified} contains magnified versions of the plots in Figure \ref{fig:simu_data_result} for three of the distributions.  From the plots, we can draw the following conclusions:
		\begin{itemize}
		\item If there is only a single cluster, then all seven algorithms produce nearly the same results. This case is essentially a classical newsvendor problem with 7,500 data observations, for which all algorithms do a good job of providing the order quantity in the test sets. 
		\item As the number of clusters increases, i.e., the number of training samples in each cluster decreases, the methods begin to differentiate somewhat. In particular:
		\item DNN-$\ell_1$, SEO, EQ, KR, and KNN perform the best and have roughly equal performance.
		\item SEO performs well when the demands are normally distributed but less well otherwise. This is because one has to assume a demand distribution in order to use SEO, and we assumed normal. If the demands happen to come from a normal distribution, therefore, SEO works well. In practice, however, the demand distribution is usually unknown and often non-normal. 
		\item EQ performs relatively well in general in this experiment because, when the data are non-noisy, it is easier to estimate a quantile. However, for both the small data set (Section~\ref{sec:toy_example_section}) and the real-world data set (Section~\ref{sec:Numerical_Experiments_Real}), which are noisier, EQ does not perform well.
		\item The performance of DNN-$\ell_2$ is quite good {\em except} in the case of normal demands with 100 or 200 clusters. In these cases, the method would benefit from further tuning (similar to the additional tuning that we did for the basket data set in Section~\ref{sec:Numerical_Experiments_Real}). 
		\item LML and RF are nearly always worse than the other methods because there is not enough data for them to learn the distribution well. (As a result, we have omitted them from Figure~\ref{fig:simu_data_result_magnified}.)
	 \end{itemize}
	 
	
\begin{figure}[]
	\centering
	\begin{subfigure}[b]{0.49\textwidth}		
		\centering
		\includegraphics[width=5.5cm]{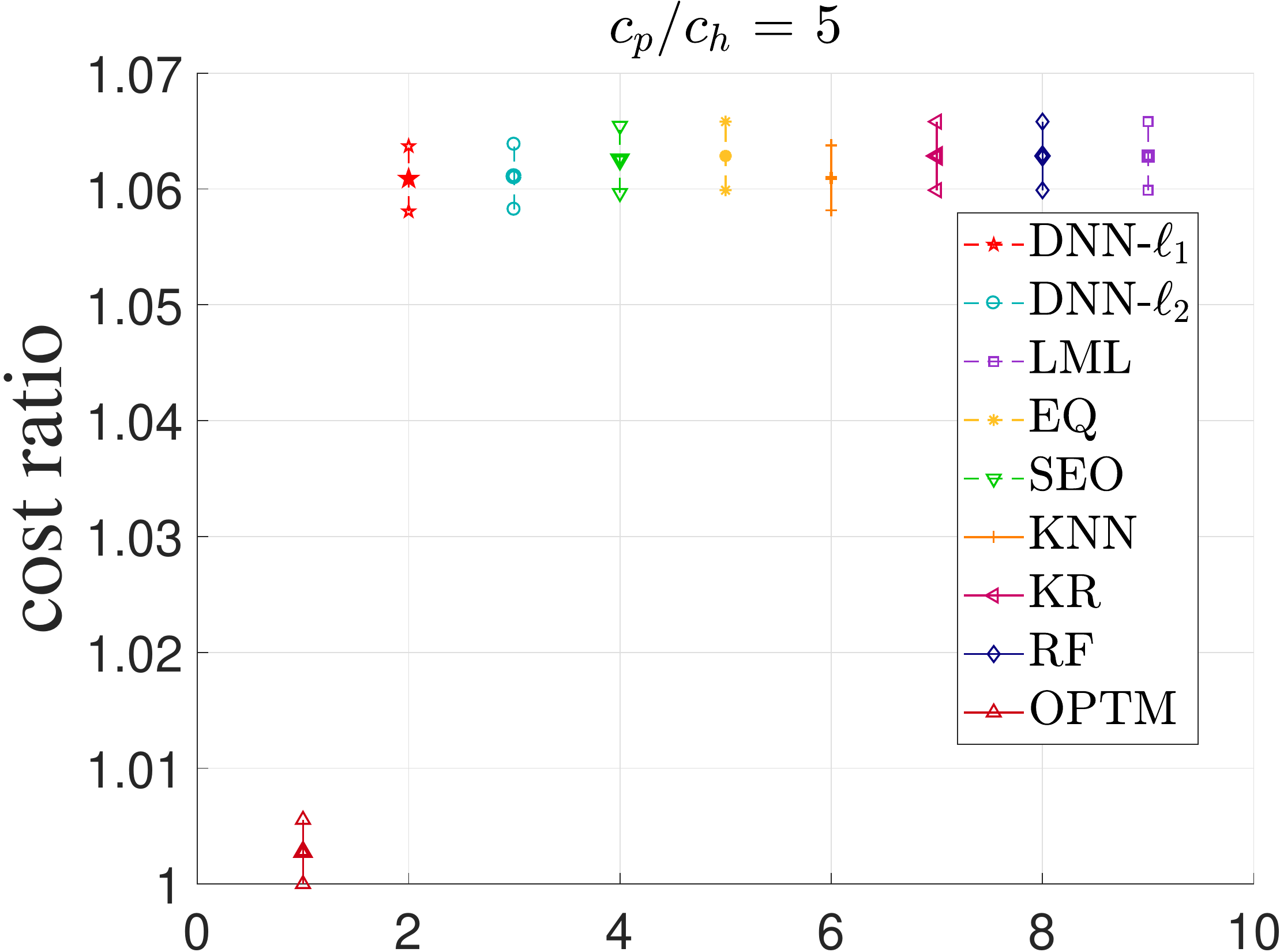}	
		\caption{1 cluster}
		\label{fig:box_normal_1}
	\end{subfigure}
	\begin{subfigure}[b]{0.49\textwidth}		
		\centering
		\includegraphics[width=5.5cm]{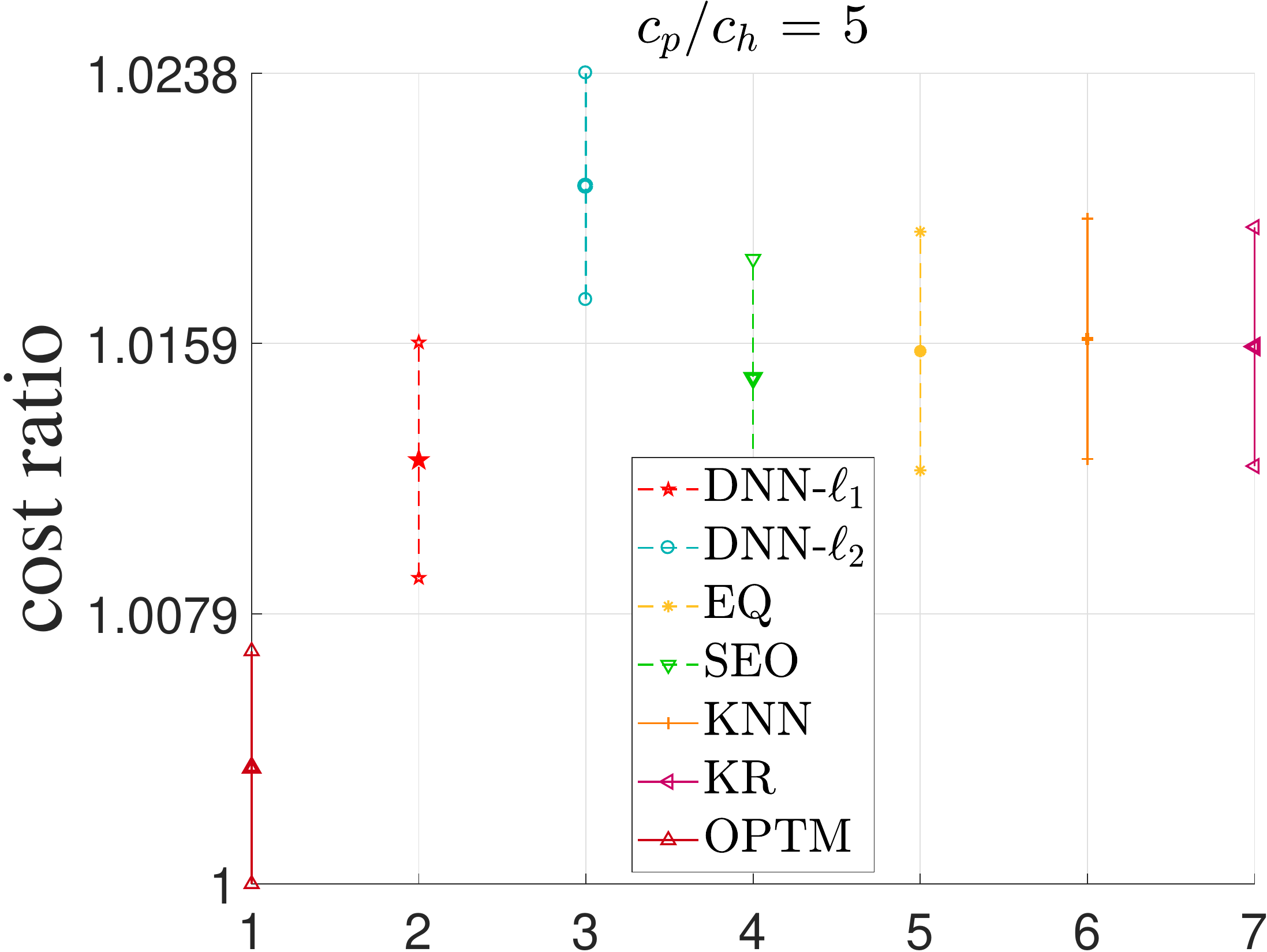}
		\caption{10 clusters}
		\label{fig:box_normal_10}
	\end{subfigure}

	\begin{subfigure}[b]{0.5\textwidth}		
		\centering		
		\includegraphics[width=5.5cm]{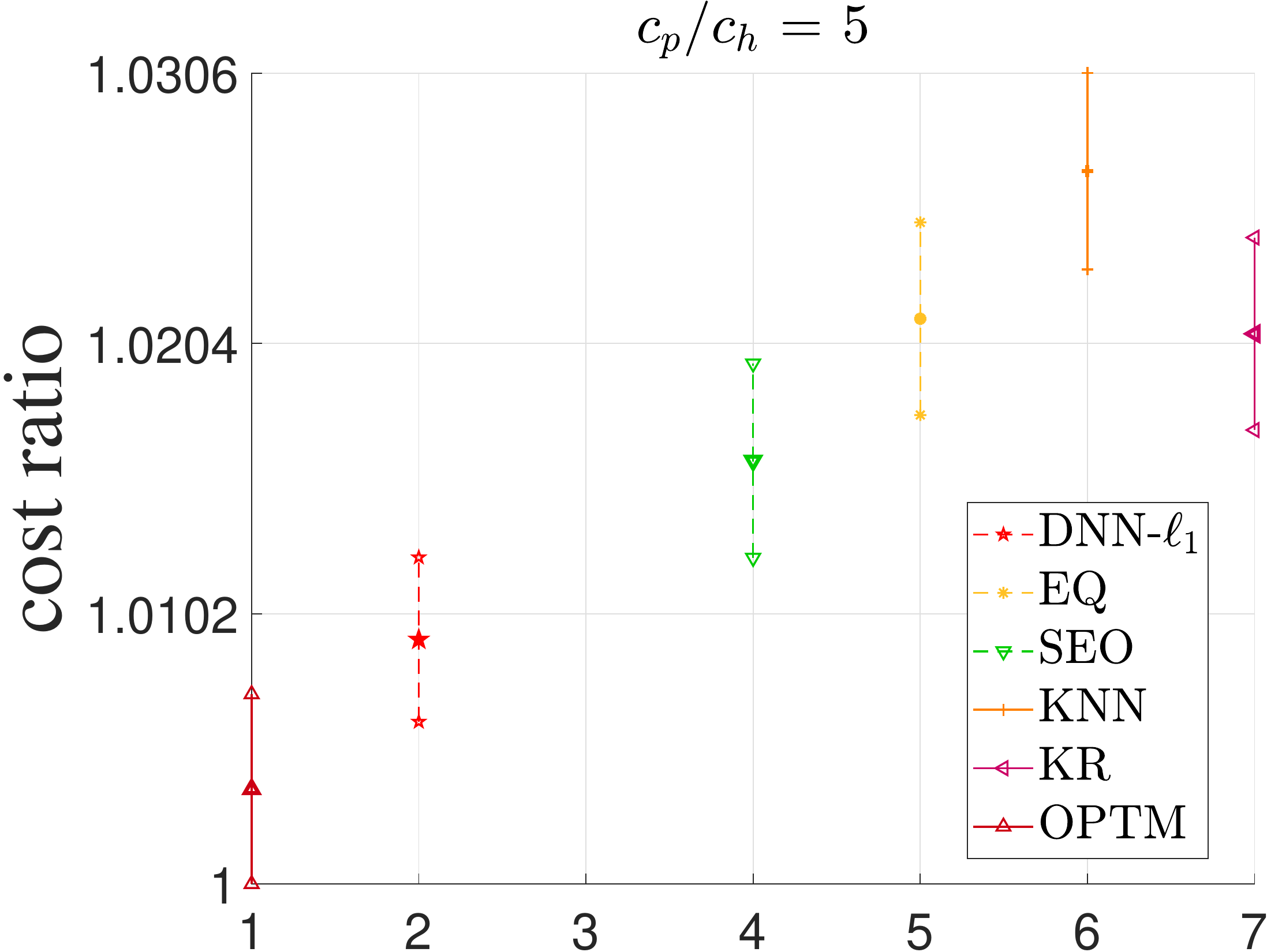}
		\caption{100 clusters}
		\label{fig:box_normal_100}			
	\end{subfigure}
	\begin{subfigure}[b]{0.49\textwidth}		
		\centering			
		\includegraphics[width=5.5cm]{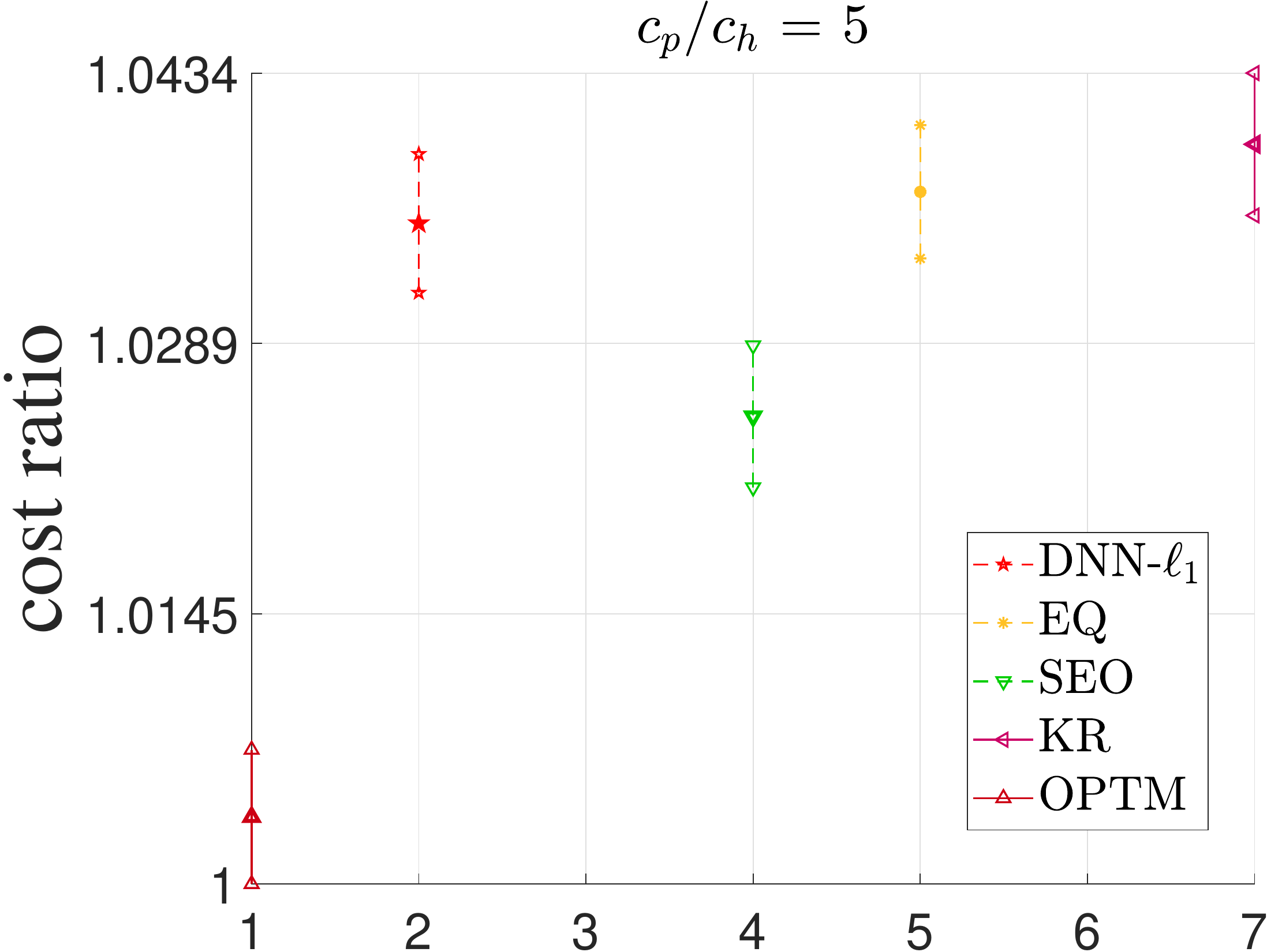}
		\caption{200 clusters}
		\label{fig:box_normal_200}			
	\end{subfigure}	
	\caption{Confidence intervals for each algorithm for normally distributed demands.}	
	\label{fig:box_normal}	
	\end{figure}

\begin{figure}[]
	\centering	
	\begin{subfigure}[b]{0.5\textwidth}		
		\centering		
		\includegraphics[width=5.5cm]{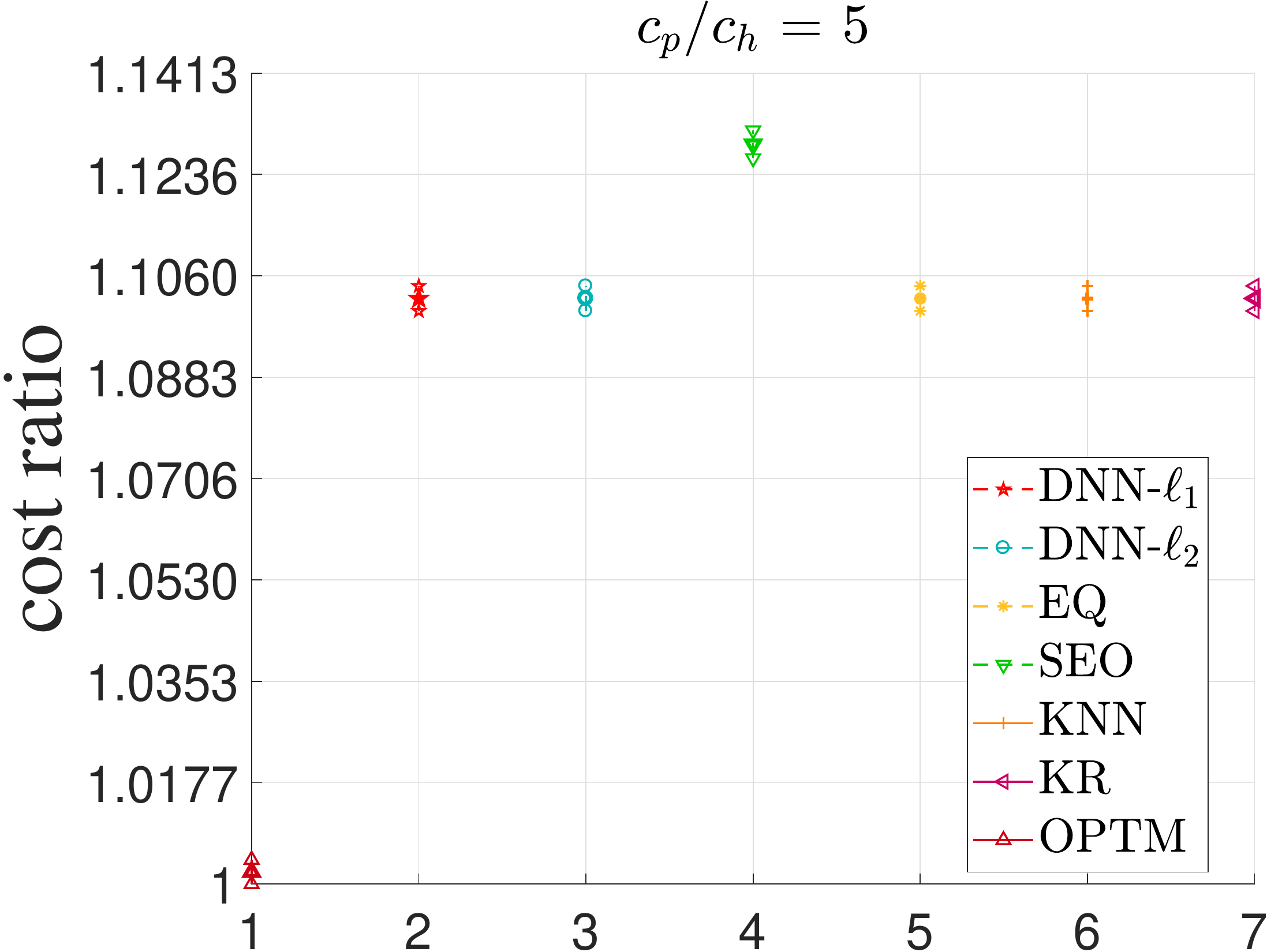}	
		\caption{1 cluster}
		\label{fig:box_uniform_1}			
	\end{subfigure}	
	\begin{subfigure}[b]{0.49\textwidth}		
		\centering			
		\includegraphics[width=5.5cm]{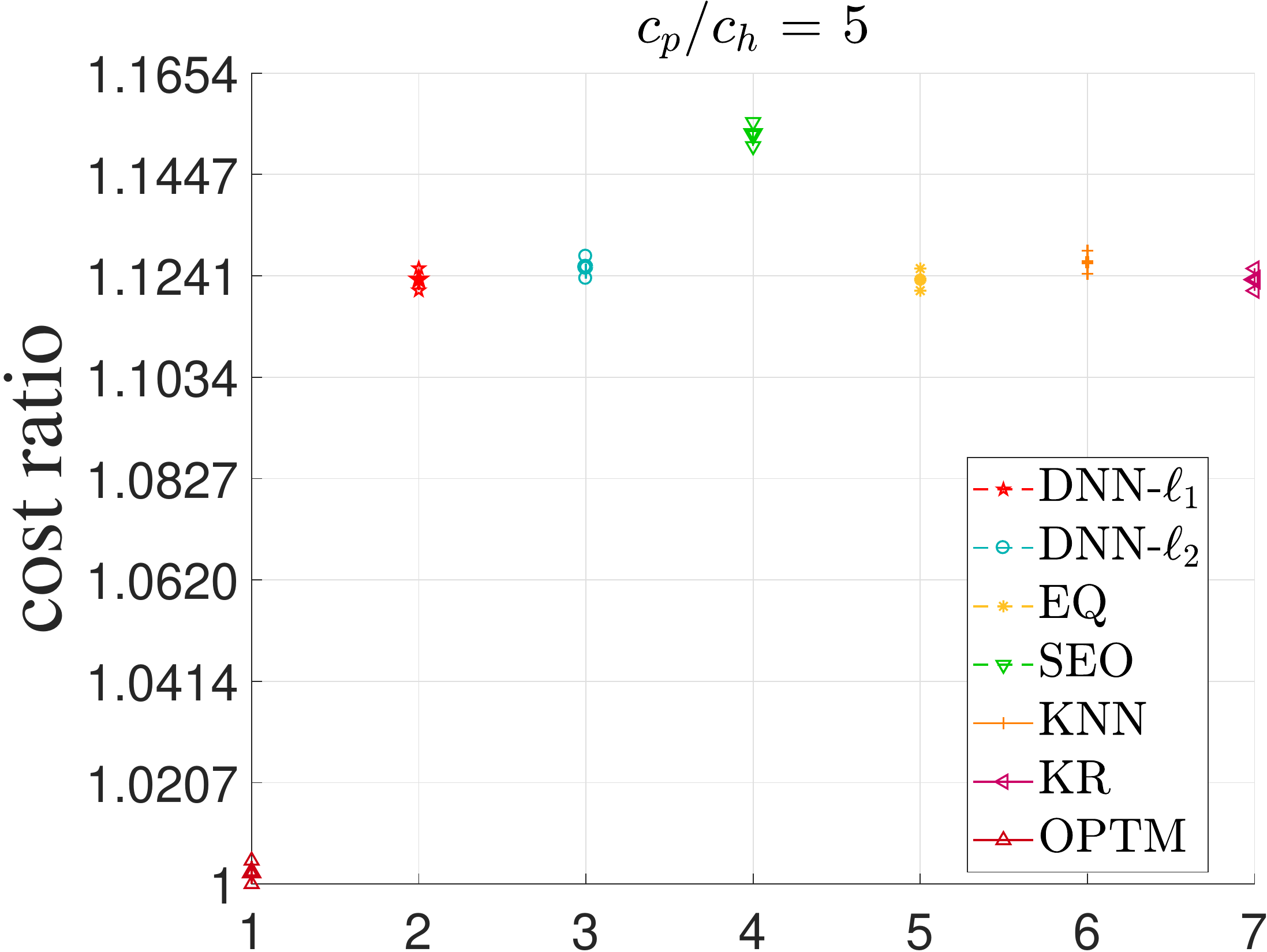}
		\caption{10 clusters}
		\label{fig:box_uniform_10}			
	\end{subfigure}	
	\begin{subfigure}[b]{0.5\textwidth}		
		\centering		
		\includegraphics[width=5.5cm]{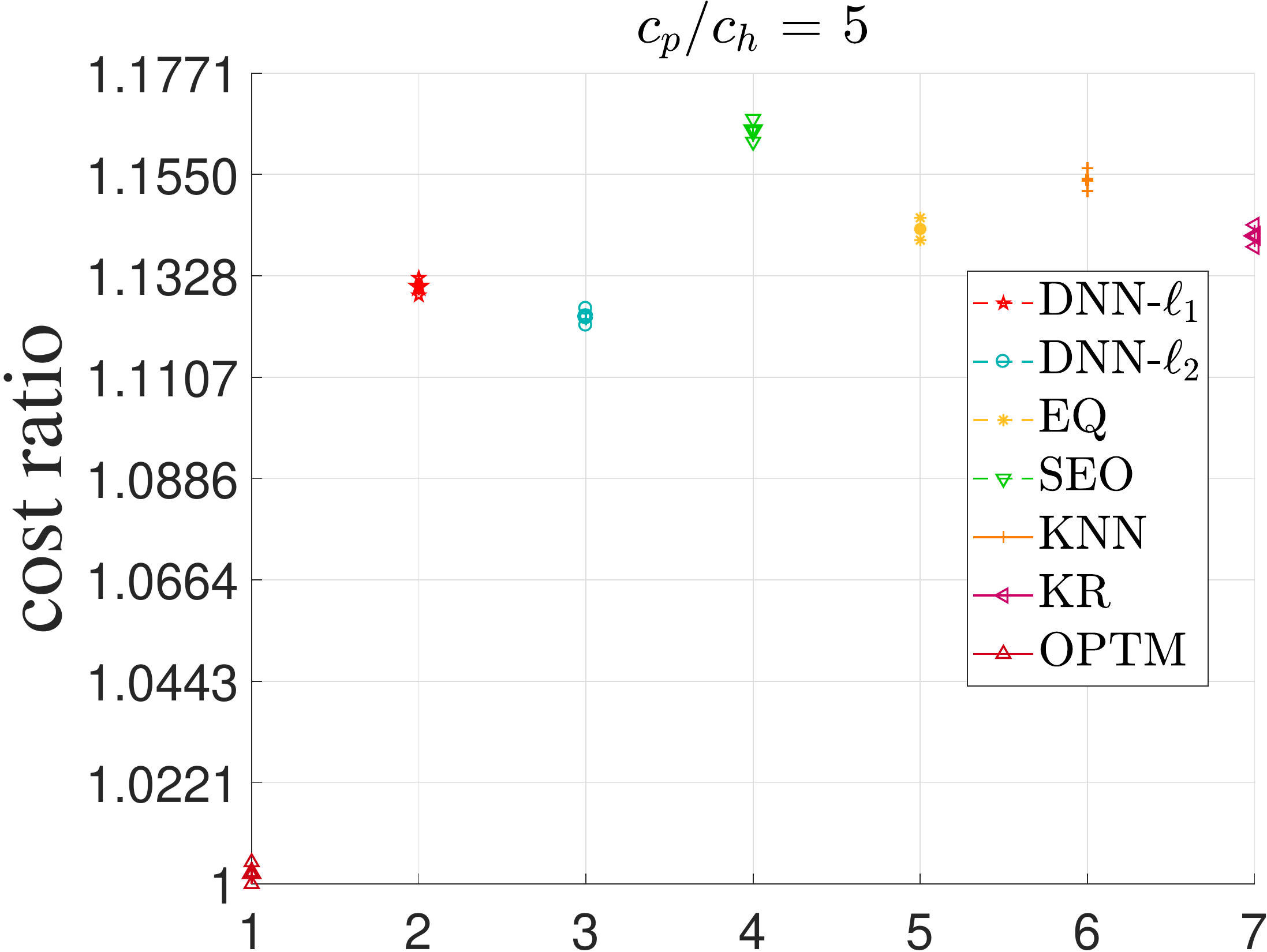}	
		\caption{100 clusters}
		\label{fig:box_uniform_100}			
	\end{subfigure}		
	\begin{subfigure}[b]{0.49\textwidth}		
		\centering		
		\includegraphics[width=5.5cm]{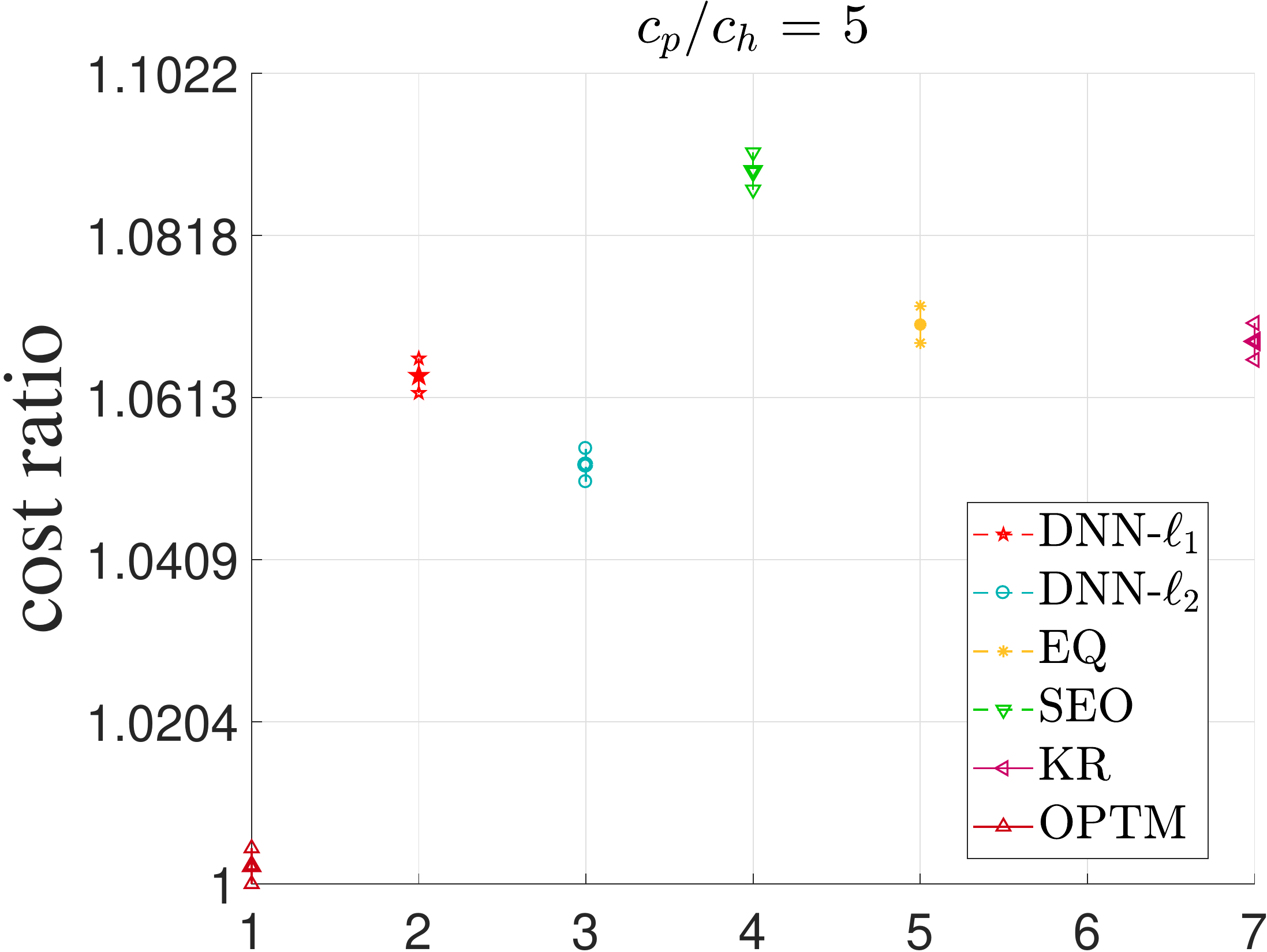}	
		\caption{200 clusters}
		\label{fig:box_uniform_200}			
	\end{subfigure}				
	\caption{Confidence intervals for each algorithm for uniformly distributed demands.}	
	\label{fig:box_uniform}	
\end{figure}

To confirm these findings statistically, Figures~\ref{fig:box_normal} and \ref{fig:box_uniform} plot 95\% confidence intervals for each algorithm for normally and uniformly distributed demands (respectively). The confidence intervals are calculated using the mean and standard error of the cost ratio over the 99 test data sets. When two confidence intervals are non-overlapping, we can conclude that the performance of the two corresponding methods is statistically different. If a given method is excluded from a plot, it means that the method is much worse than the methods that are plotted. From these figures, we can draw the following conclusions:
	\begin{itemize}
	\item DNN-$\ell_1$ is statistically better than all other methods for some cases (e.g., normal demands with 200 clusters); is in statistical second place to SEO for normal demands with 200 clusters and to DNN-$\ell_2$ for uniform demands with 100 and 200 clusters; and is tied for first place in all other cases. 
	\item SEO is statistically better than all other methods for normal demands with 200 clusters and statistically worse than all other methods for uniform demands with any number of clusters. It is tied with other methods for most other instances.
	\item DNN-$\ell_2$, EQ, KNN, and KR are, in most cases, in a statistical tie.
	\item LML and RF are statistically worse than all other methods, except in the case of normal demands with 1 cluster. 
	\item In nearly every instance, no method obtains solutions that are statistically equal to the optimal solution. The exception is normal demands with 100 clusters, for which DNN-$\ell_1$ is statistically tied with the optimal solution.
	\end{itemize}

\begin{figure}
	\centering
		\includegraphics[scale=0.5]{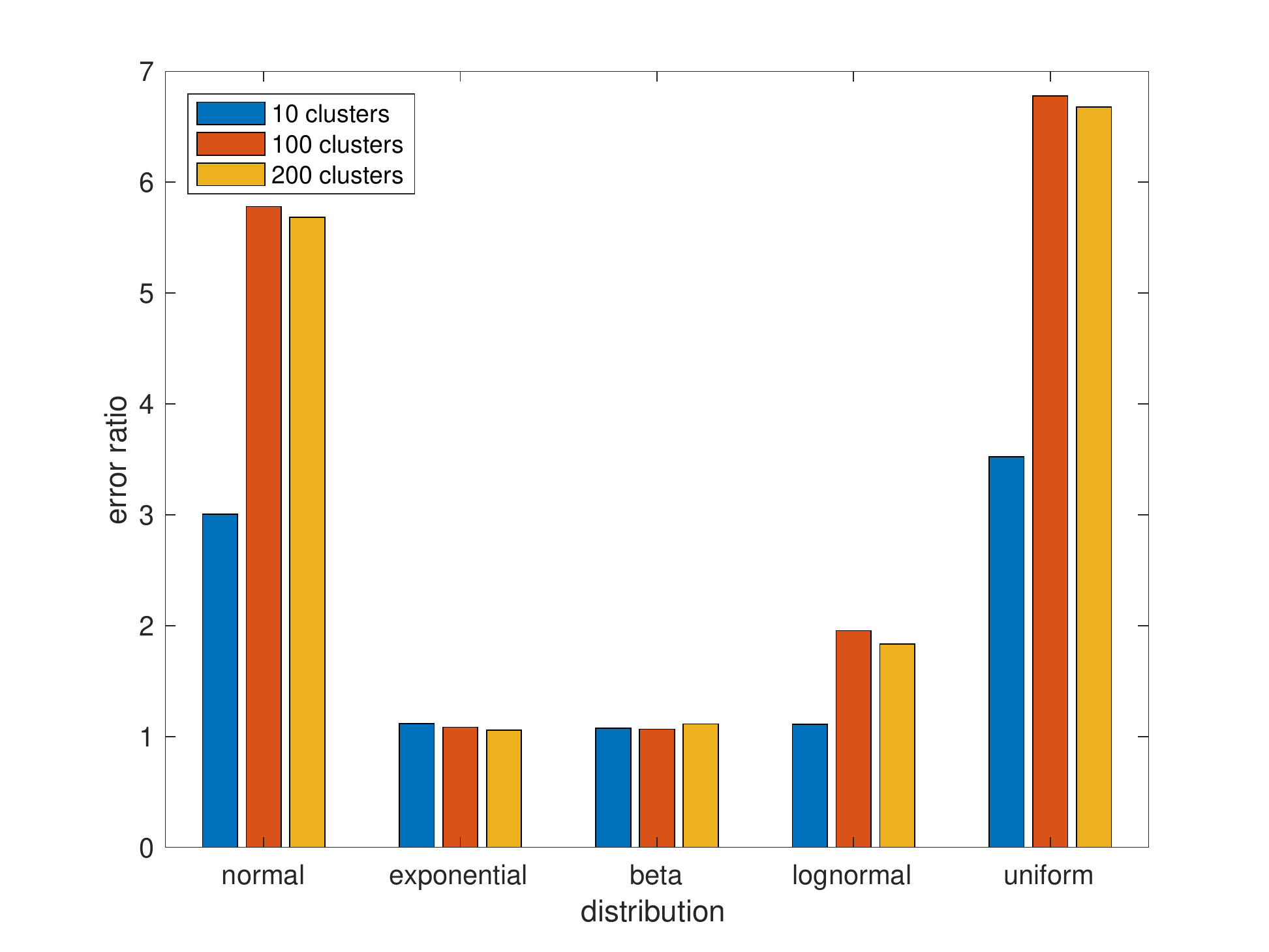}
		\caption{Error ratio from ignoring clusters when solving MFNV.}
	\label{fig:ignore_clusters}
	\end{figure}

Suppose we take a naive approach toward the MFNV problem and ignore the data features, optimizing the inventory level as though there were only a single cluster. How significant an error is this? To answer this question, we solved the problem using DNN-$\ell_1$, grouping all of the data into a single cluster. (Note that this data set is different from the 1-cluster data sets discussed above. The data sets above assume there {\em is} only a single cluster, i.e., all demand records have identical feature values, whereas the data set here has multiple sets of feature values, but we are ignoring them to emulate the naive approach.) Figure~\ref{fig:ignore_clusters} plots the ratio between the cost of the resulting solution and the cost of the DNN-$\ell_1$ solution that accounts for the clusters, for the five probability distributions and for data sets with 10, 100, and 200 clusters. Clearly, the error resulting from this naive approach can be significant: They range from 5.6\% (for the exponential distribution with 200 clusters) to 677.9\% (for the uniform distribution with 100 clusters). In general these errors will change with the probability distributions and their parameters, but it is clear that it is important to consider clusters when faced with featured data, and costly to ignore them. 

\subsection{Numerical Results: Summary}

	Our recommendations for which method to use are as follows. If the data set is noisy, like most real-world data sets, our experiments show that DNN is the most reliable algorithm, with the caveat that careful hyperparameter tuning is required. 
	If the data are non-noisy (they come from a single probability distribution) and the number of historical samples is small (say, fewer than 10 records per combination of features), DNN tends to outperform the other methods. As the number of historical records begin to increase, either EQ, SEO, DNN, KR, RF, or KNN is a reasonable choice. Finally, if there are a large number of non-noisy historical demand records for each combination of features (say, at least 10,000), then the algorithms all work roughly equally well, and it may be best to choose EQ or SEO, since they do not need any hyperparameter tuning.

%

	\section{Extension to $(r,Q)$ Policy}\label{sec:r_q}

	In this section, we extend our DNN approach to optimize the parameters of an $(r,Q)$ inventory policy, in order to demonstrate that the method can be adapted to other inventory problems, and especially to problems that cannot be solved simply by estimating the quantile of a probability distribution. 
	Consider a continuous-review inventory optimization problem with stochastic demand, such that the mean of demand per unit time is $\lambda$. 
	Placing an order incurs a fixed cost $K$, and the order arrives after a deterministic lead time of $L\ge0$ time units. Unmet demand is backordered. We assume the firm follows an $(r,Q)$ inventory policy: Whenever the inventory position falls to $r$, an order of size $Q$ is placed. The aim of the optimization problem is to determine $r$ and $Q$.  
	
	If we know the true demand distribution, the optimal $r$ and $Q$ can be obtained by solving a a convex optimization problem; see \cite{hadley1963analysis} or \cite{Zh92}. However, heuristic approaches are commonly used to obtain approximate values for $r$ and $Q$; for a discussion of these, see \cite{snyder2018fundamentals}. We use the so-called expected-inventory level (EIL) approximation, which is arguably the most common approximation for the $(r,Q)$ optimization problem. 
	The EIL approximates the expected cost function as
	\begin{equation}
	g(r,Q) = c_h \left(r - \lambda L + \frac{Q}{2} \right) + \frac{K\lambda}{Q} + \frac{c_p\lambda n(r)}{Q},
	\label{eq:eil_rQ_cost_function}
	\end{equation}
	where 
	\begin{equation*}
	n(r) = \int_{r}^{\infty} (d-r)f(d)dd
	\end{equation*}
	and $f(d)$ is the demand distribution. The cost function \eqref{eq:eil_rQ_cost_function} can be optimized through an iterative algorithm proposed by  \cite{hadley1963analysis}, again assuming that the demand distribution is known.
	
	Of course, in practice, the demand distribution is often not known, which is where DNN becomes a useful approach. 
	In order to use DNN to obtain the policy parameters, we propose a DNN network similar to that used for the newsvendor problem, except that it has two outputs,  $r$ and $Q$.
	We use the cost function \eqref{eq:eil_rQ_cost_function} as the loss function for the DNN, and in place of $n(r)$ we use the unbiased estimator $\frac1m \sum_{i=1}^m (d_i-r_i)^+$. 
	In addition, in order to avoid negative values for $r$ and $Q$, we use $r^+$ and $Q^+$ in the DNN loss function, and also add a penalty for negative values of $r$ and $Q$ into the DNN loss function:
	\begin{equation*}
	l(r,Q) = c_h \left(r^+ - \lambda L + \frac{Q^+}{2} \right) + \frac{K\lambda}{Q^+} + \frac{c_p\lambda n\left(r^+\right)}{Q^+} + \eta_Q Q^- + \eta_r r^-,
	\end{equation*}
	where $\eta_r$ and $\eta_Q$ are the penalty coefficients for negative $r$ and $Q$, respectively.

	\subsection{Numerical Experiments}\label{sec:r_q_numerical_experiments}
	
	In order to see the effectiveness of the proposed algorithm for the $(r,Q)$ optimization problem,
	we tested both algorithms on a problem with $K=2$, $L=1$, $\lambda=10$, $c_p=2$, and $c_h=1$.
	We used the iterative algorithm by \cite{hadley1963analysis} to obtain the optimal $r$ and $Q$ that minimize \eqref{eq:eil_rQ_cost_function}. (We will refer to this as the {\em EIL algorithm}.) Since the algorithm needs the demand distribution, similar to the approach in Section \ref{sec:Numerical_experiments_simulation}, we fit a normal distribution to each cluster and use it to obtain $(r,Q)$ for that corresponding cluster. 	
	
	When testing the DNN algorithm on this problem, we did not perform any hyper-parameter tuning; all instances use the same hyper-parameters. 
	For most instances we used a DNN network of shape $[43, 90, 100, 56, 2]$ in which all nodes have a {\tt Relu} activation function, where {\tt Relu}$(x) = x^+$. The only exception is that in the case of uniform distributed data with 10 clusters, the algorithm did not converge, so we instead used a smaller network of shape $[43, 90, 56, 2]$. 
	We used the Adam optimizer \citep{kingma2014adam} to optimize the weights of the network with learning rate$=0.001$, $\beta_1 = 0.9$, $\beta_2=0.999$, $\epsilon=1e-8$, a batch size of $128$, and $\gamma=0.009$. 
			
	In what follows, we demonstrate the results of both algorithms on two datasets that we used when testing the newsvendor problem: the basket data set, which is presented in Section \ref{sec:r_q_numerical_experiments_basket}, and the five randomly generated datasets, presented in Section \ref{sec:r_q_numerical_experiments_random}.

	\subsubsection{Basket Dataset}\label{sec:r_q_numerical_experiments_basket}

	We obtained $(r,Q)$ values using both algorithms. The solution found by the EIL algorithm incurs a cost of 121,772, while that obtained by DNN has a cost of 117,538, 3.5\% better than EIL. At first this may seem surprising, since the EIL algorithm is an exact algorithm to optimize the cost function \eqref{eq:eil_rQ_cost_function} (though of course \eqref{eq:eil_rQ_cost_function} is itself an approximation of the exact cost function). However, recall that the basket dataset is noisy and contains few historical observations (between 1 and 9) per cluster, but the EIL algorithm assumes the demands are normally distributed. This assumption is inaccurate for the basket dataset. On the other hand, DNN considers the feature values and in three epochs optimizes the weights of the network, and in doing so is able to learn better $(r,Q)$ values to minimize the objective.
			
	\subsubsection{Randomly Generated Data}\label{sec:r_q_numerical_experiments_random}
	
\begin{figure}
	\centering
		\includegraphics[scale=0.25]{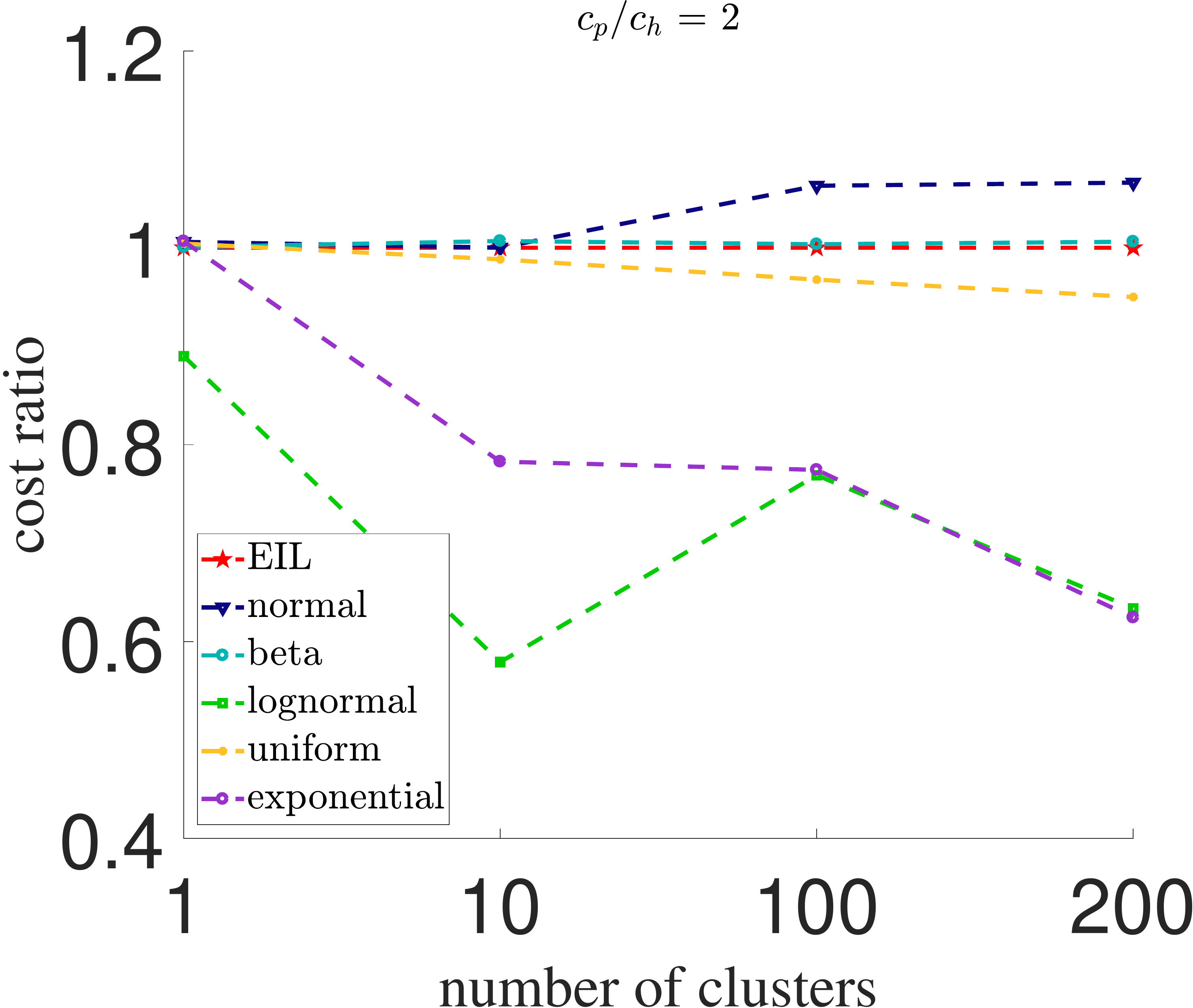}
		\caption{The results for randomly generated datasets for the $(r,Q)$ model.}
	\label{fig:r_q_simulation_2_1}
	\end{figure}

	In order to further explore the performance of both algorithms, we tested their performance on the randomly generated datasets in Section \ref{sec:Numerical_experiments_simulation}. 
	Just as in the newsvendor problem, we assume we do not know the demand distribution and instead approximate a normal distribution in each cluster to obtain the solution using EIL.
	Under DNN, each problem ran for 50 epochs and all of them converged after at most 10 epochs of training. 
	The results of all demand distributions are shown in Figure \ref{fig:r_q_simulation_2_1}. 
	As shown in the figure, when the data are generated from a normal distribution, EIL finds the optimal solution but the  DNN solution is close, with around a 3\% gap, on average, for four clusters. 
	DNN provides a near-0\% gap for the beta distribution, and in all other cases DNN provides a better solution than EIL, with an average cost ratio of $0.83$. 
	
	Let us more closely examine one instance, the normally distributed dataset, for which the EIL solution is optimal. When there is only one cluster, the optimal solution from EIL is $(r,Q) = (0.00,45.16)$, whereas DNN obtains $(r,Q)=(0.38,45.01)$, which is quite close.
	Similarly, when there are 10 clusters, the DNN $(r,Q)$ is quite close to the optimal solutions, as  shown in Table \ref{tb:r_q_value_comparison_normal_10}. 
	As a result, the costs of the solutions obtained by the two algorithms are almost equal. 
	Similar results also emerge from the instances with 100 and 200 clusters.
	
	\linespread{1}

	\begin{table}
		\centering
		\caption{EIL and DNN values of  $(r,Q$) for the normally distributed dataset with 10 clusters.}
		\label{tb:r_q_value_comparison_normal_10}
		\begin{tabular}{l|llllllllll}
			Cluster & 1 & 2 & 3& 4& 5& 6& 7& 8& 9& 10 \\ \hline 
			Optimal $r$ &  0.00 & 0.00 & 0.00 & 0.00 & 0.00 & 0.00 & 0.00 & 0.00 & 0.00 & 0.00  \\ 
			DNN $r$& 0.18& 0.14& 0.21& 0.17& 0.18& 0.1& 0.13& 0.2& 0.19& 0.10 \\ \hline
			Optimal $Q$ & 118.49& 77.72& 141.56& 100.20& 109.73& 63.56& 89.67& 134.31& 126.65& 45.17	\\
			DNN $Q$& 119.25& 77.98& 141.00& 102.00 & 109.82& 63.92& 89.26& 134.35& 127.35& 44.89 \\
			\hline
		\end{tabular}
	\end{table}

	\linespread{2}
	
	To summarize, if the true distribution is available, our DNN method and the classical EIL approach work almost equally well. 
	However, EIL's performance deteriorates when the true demand distribution is not known, even if there is a relatively large amount of historical data. In contrast, DNN works well when the true demand distribution is unknown, even if the historical dataset is small and/or noisy. 

	\section{Conclusion}
	\label{sec:conclusion}
	
	In this paper, we consider the multi-feature newsvendor (MFNV) problem. 
	If the  probability distribution of the demands is known for every possible combination of the data features, there is an exact solution for this problem. However, approximating a probability distribution is not easy and produces errors; therefore, the solution of the newsvendor problem also may be not optimal. Moreover, other approaches from the literature do not work well when the historical data are scant and/or volatile. 
	
	To address this issue, we propose an algorithm based on deep learning to solve the MFNV. 
	The algorithm does not require knowledge of the demand probability distribution and uses only historical data. 
	Furthermore, it integrates parameter estimation and inventory optimization, rather than solving them separately.
	Extensive numerical experiments on real-world and random data demonstrate the conditions under which our algorithm works well compared to the algorithms in the literature. 
	The results suggest that when the volatility of the demand is high, which is common in real-world datasets, deep learning works very well.
	When the data can be represented by a well-defined probability distribution, in the presence of enough training data, a number of approaches, including DNN, have roughly equivalent performance. 
	
	Furthermore, we extend our DNN approach to the $(r,Q)$ inventory optimization problem, to demonstrate that our approach is applicable in more general settings, especially those that cannot be solved by estimating a quantile. Our computational results show that the DNN approach works well when the historical data are noisy and/or sparse, and that it often outperforms the ``exact'' algorithm when the true demand distribution is unknown (since the exact algorithm must make an assumption about the distribution). 

	Motivated by the results of deep learning on both newsvendor and $(r,Q)$ problems, we suggest that this idea can be extended to  other supply chain problems. For example, since general multi-echelon inventory optimization problems are very difficult, deep learning may be a good candidate for solving these problems. 
	Another direction for future work could be applying other machine learning algorithms to exploit the available data in the newsvendor problem.
	
	
	\small
		
	\bibliographystyle{myplainnat}
	\bibliography{newsvendor}

	\normalsize 
	\clearpage
	\appendix
	
	\section{Proofs of Propositions~\ref{prop:gradient1} and \ref{prop:gradient2}}
	\label{sec:proofs}
	
		These proofs are based on the general idea of the back-propagation algorithm and the way it builds the gradients of the network. For further details, see \cite{lecun2015deep}.

	
	%
	%
	{\bf Proof of Proposition~\ref{prop:gradient1}.}  
	%
	%
	%
	%
	To determine the gradient with respect to the weights of the network, we first consider the last layer, $L$, which in our network contains only one node. Note that in layer $L$, $y_i^q = a_1^L$. 
	So, we first obtain the gradient with respect to $w_{j1}$, which connects node $j$ in layer $L-1$ to the single node in layer $L$, and then recursively calculate the gradient with respect to other nodes in other layers. 
	
	First, consider the case of excess inventory ($d_i^q \le y_i^q$). Recall from \eqref{eq:delta} that $\delta_j^l=\frac{\partial E_i^q }{\partial a_j^l} ({g}^l_j)' (z_j^l)$. Then $\delta_1^L = c_h ({g}^L_1)' (z_1^L)$, since $E_i^q = c_h(a_1^L - d_i^q)$. Then:
	\begin{equation}	
	\begin{split}
	\label{eq:gradient_excess_nw}	
	\frac{\partial E_i^q}{\partial w_{j1}} & = c_h \frac{\partial (y_i^q - d_i^q) }{\partial w_{j1}} \\
	& = c_h \frac{\partial a^L_1  }{\partial w_{j1}} \quad \text{(since $d_i^q$ is independent of $w_{j1}$)} \\	
	& = c_h \frac{\partial g^L_1(z^L_1)  }{\partial w_{j1}} \\
	& = c_h  \frac{\partial g^L_1(z^L_1) }{\partial z^L_1} \frac{\partial z^L_1 }{\partial w_{j1}} \quad \text{(by the chain rule)}  \\
	&= c_h ({g}^L_1)' (z^L_1) a_j^{L-1} \quad \text{(by \eqref{eq:alj})} \\
	&= \delta_1^L(h) a_j^{L-1}  \quad \text{(by \eqref{eq:deltaph})}.
	\end{split} 
	\end{equation}	
	%
	
	Now, consider an arbitrary layer $l$ and the weight $w_{jk}$ that connects node $j$ in layer $l$ and node $k$ in layer $l+1$. 
	Our goal is to derive $\delta_j^l = \frac{\partial E_i^q }{\partial z_j^l}$, from which one can easily obtain $\frac{\partial E_i^q }{\partial w_{jk}}$, since 
	\begin{equation}
	\label{eq:gradient_delta}
		\frac{\partial E_i^q }{\partial w_{jk}} = \frac{\partial E_i^q }{\partial z_j^l} \frac{\partial z_j^l }{\partial w_{jk}} 
		= \delta_j^l a_j^l \\
	\end{equation}
	using similar logic as in \eqref{eq:gradient_excess_nw}.
	To do so, we establish the relationship between $\delta_j^l$ and $\delta_k^{l+1}$. 
	\begin{equation}
	\label{eq:deltas_connection}
	\begin{split}
		\delta_j^l & = \frac{\partial E_i^q }{\partial z_j^l} \\
		& = \sum_{k} \frac{\partial E_i^q }{\partial z_k^{l+1}} \frac{\partial z_k^{l+1} }{\partial z_j^l} \\
		& = \sum_{k} \delta_k^{l+1} \frac{\partial z_k^{l+1} }{\partial z_j^l} 
	\end{split}
	\end{equation}
	Also, from \eqref{eq:zjl}, we have 
	\begin{equation*}
	\begin{split}
		z_{k}^{l+1} & = \sum_j w_{jk}a_j^l  
		 = \sum_j w_{jk} g_j^l(z_j^l) \\ 
	\end{split}
	\end{equation*}
	Therefore, 
	\begin{equation}
	\label{eq:derivative_of_z_wrt_z}
	\begin{split}
	\frac{\partial z_k^{l+1} }{\partial z_j^l}  & = w_{jk} ({g}^l_j)'(z_j^l). 
	\end{split}
	\end{equation}
	Plugging \eqref{eq:derivative_of_z_wrt_z} into \eqref{eq:deltas_connection}, results in \eqref{eq:loss_and_delta}.
	\begin{equation}
		\label{eq:loss_and_delta}
		\delta_j^l = \sum_{k} w_{jk} \delta_k^{l+1} ({g}^l_j)'(z_j^l).
	\end{equation}
	We have now calculated $\delta_j^l$ for all $l=1,\ldots,L$ and $j=1,\ldots,nn_l$. 
Then, substituting \eqref{eq:loss_and_delta} in \eqref{eq:gradient_delta}, the gradient with respect to any weight of the network is:
	\begin{equation}
	\label{eq:final_gradient}
	\frac{\partial E_i^q }{\partial w_{jk}} 
	= a_j^l \sum_{k} w_{jk} \delta_k^{l+1} {g}'^l_j(z_j^l).
	\end{equation}
	%

	Similarly, for the shortage case in layer $L$, we have:
	\begin{equation}	
	\begin{split}
	\label{eq:gradient_shortage_nw}
	\frac{\partial E_i^q}{\partial w_{j1}} & = - c_p \frac{\partial (d_i^q - y_i^q) }{\partial w_{j1}} \\
	& = c_p \frac{\partial (a^L_1 ) }{\partial w_{j1}} \\	
	& = c_p \frac{\partial (g^L_1(z^L_1) ) }{\partial w_{j1}} \\
	& = c_p  \frac{\partial (g^L_1(z^L_1)) }{\partial z^L_1} \frac{\partial (z^L_1) }{\partial w_{j1}} \\
	&= c_p a_j^{L-1} ({g}^L_1)' (z^L_1) \\
	&= \delta_1^L(p)a_j^{L-1}.
	\end{split}
	\end{equation}		
	Using the chain rule and following same procedure as in the case of excess inventory, the gradient with respect to any weight of the network can be obtained. Summing up \eqref{eq:gradient_excess_nw}, \eqref{eq:final_gradient} and \eqref{eq:gradient_shortage_nw}, the gradient with respect to the $w_{jk}$ is:
	\begin{equation*} 	
	\frac{\partial E_i^q}{\partial w_{jk}} = \begin{cases}	
	a_j^l \delta_j^l(p) & \text{if } y_i^q < d_i^q, \\ 
	a_j^l \delta_j^l(h) & \text{if } d_i^q \le y_i^q.
	\end{cases}
	\end{equation*} 	    
	\qed

	{\bf Proof of Proposition~\ref{prop:gradient2}.} Consider the proposed revised Euclidean loss function 
	defined in \eqref{quadratic_nw}. 
	Using similar logic as in the proof of Proposition~\ref{prop:gradient1}, we get that the gradient of the loss function at the single node in layer $L$ is 
	\begin{equation}	
	\begin{split}
	\label{gradient_excess}	
	\frac{\partial E_i^q}{\partial w_{j1}} & = c_h( y_i^q - d_i^q) \frac{\partial (y_i^q - d_i^q) }{\partial w_{j1}} \\
	&= ( y_i^q - d_i^q) a_j^{L-1} \delta_1^L(h).
	\end{split} 
	\end{equation}	
	in the case of excess inventory and 
	\begin{equation}	
	\begin{split}
	\label{gradient_shortage}
	\frac{\partial E_i^q}{\partial w_{j1}} & = - c_p (d_i^q - y_i^q) \frac{\partial (d_i^q - y_i^q) }{\partial w_{j1}} \\
	&= (d_i^q - y_i^q) a_j^{L-1} \delta_1^L(p).
	\end{split}
	\end{equation}		
	in the shortage case. Again following the same logic as in the proof of Proposition~\ref{prop:gradient1}, the gradient with respect to any weight of the network can be obtained:
	\begin{align*}
	\frac{\partial E_i^q}{\partial w_{jk}} &=  \left\{\begin{matrix}
	( d_i^q - y_i^q) a^l_j \delta_1^l(p)  & \text{if } y_i^q < d_i^q\\ 
	( y_i^q - d_i^q) a^l_j \delta_1^l(h)  & \text{if } d_i^q \le y_i^q.
	\end{matrix}\right. 
	\end{align*} 		
	\qed

	\section{Grid Search for Basket Dataset}\label{appendix basket_grid_search}
	
	In this appendix, we discuss our method for performing a more thorough tuning of the network for DNN-$\ell_2$, as discussed in Section~\ref{sec:Numerical_Experiments_Real}. 
	We used a large, two-layer network with 350 and 100 nodes in the first and second layer, respectively. In order to find the best set of parameters for this model, a grid search is used. We considered three parameters,  $lr$, $\lambda$, and $\gamma$; $\lambda$ is the regularization coefficient, and $lr$ and $\gamma$ are parameters used to set the learning rate. In particular, we set $lr_t$, the learning rate used in iteration $t$, using the following formula:
	\begin{equation*}
		lr_t = lr \times (1 + \gamma \times t)^{-0.75}.
	\end{equation*}
	We considered parameter values from the following sets:
	\begin{equation*}
	\begin{array}{lll}	
	\gamma \in & \{0.01, 0.005, 0.001, 0.0001, 0.0005, 0.00005\} \\
	\lambda \in & \{0.01, 0.005, 0.001, 0.0005, 0.0001, 0.00005\} \\
	\text{lr} \in & \{0.001, 0.005, 0.0005, 0.0001, 0.00005, 0.00003, 0.00001, 0.000009, 0.000008, 0.000005 \},  
	\end{array}		
	\end{equation*} 	
	The best set of parameters among these 360 sets were $\gamma = 0.00005, \lambda = 0.00005$, and $lr=0.000009$. These parameters were used to test integer values of $c_p/c_h \in \{3,\ldots,9\}$ in Figure \ref{fig:real_data_resultx}, for the series labeled DNN-$\ell_2$-T.

	\section{A Tuning-Free Neural Network to Solve the Newsvendor Problem}\label{sec:appnd:tune-free-structure}
	
	To tune the hyper-parameters of the DNN in Section~\ref{sec:Numerical Experiments}, we used an extension of the random search algorithm  \citep{bergstra2012random} called HyperBand \citep{li2016hyperband}---in particular, to determine the network structure, learning rate, and regularization coefficient. 
	However, a user of our model might not have the time, resources, or expertise to follow a similar procedure. 
	Even cheaper procedures like Bayesian optimization \citep{snoek2012practical, gardner2014bayesian} are still too time consuming and too complex to implement. To address this issue, in this section we propose a computationally cheap approach to set up a network structure without extensive tuning. Our approach  provides quite good results on a wide range of problem parameters.
	
	The network structure should have a direct relation with the number of training samples $n$, the number of features $p$, and the range $r_i$ that feature $f_i$, $i=1,\dots,p$, can take values from. 
	For example, a feature $f_i$ which represents the day of week takes values between 1 and 7, and the one-hot-encoded version is a categorical feature with 7 categories; so, $r_i = 7$. For a continuous feature like the sales quantity, $r_i$ may be an interval such as $[0,\infty]$.
	These characteristics---the number of features and the range of values for each feature---affect both the number of layers in the network and the numbers of nodes in each layer. 
	For instance, if the number of features is small and the features take on only a few values, a trained DNN returns a solution that minimizes the average loss value. In this case a small network can provide quite good results. 
	On the other hand, when the number of features is relatively large and each feature can take values from a large range or set, the DNN must be able to distinguish among a large number of cases. In this event, the DNN network must be relatively large.
	
	Now, consider the newsvendor problem with $p$ features. 	
	In the datasets that we considered, the features are quite simple, e.g., receipt date and item category. However, we wish to propose a general structure for prospective users of our model, so we assume one may use more complex features, either categorical or continuous. (However, we assume the input cannot be an image, so we do not need a convolutional \cite{goodfellow2016deep} network.) 
	Thus, we propose a three-layer network in which the number of nodes in the first, second, and third hidden layers equal $aq$, $bq$, and $cq$, respectively, where  $a$, $b$, and $c$ are constants (by default we use  $a=1.5$, $b=1$, and $c=0.5$), and where  $q$ is defined as follows. Let $q_v$ be the number of continuous features, let $P_c \subseteq \{1,\ldots,p\}$ be the set of categorical features, and let
	\begin{equation*}
	q_u = \min \left\{\sum_{i\in P_c} r_i, \prod_{i\in P_c} r_i \right\}.
	\end{equation*}
	In words, $q_u$ is the smallest number that can represent all combinations of categories. Let $q = q_u + q_v$. Finally, the number of input nodes also equals $q$, and the output layer includes a single node.

	Using this approach, if the number of features is small, the number of DNN weights to optimize is small, and if the number of features is large, the number of weights is large. 
	Using the default values of $a, b$, and $c$, the proposed network has $m=\frac12 q(7q+1)$ weights, which should be smaller than the number of training records. 
	If $m>n$, there is a chance of over-fitting, and if $m\gg n$, the DNN over-fits the training data with high probability, in which case the number of DNN variables must be reduced. 
	In this case one should select smaller values of the coefficients $a$, $b$, and $c$ to reduce the number of nodes in each layer.
	Finally, using the default coefficient values, the resulting network has size $[q, 1.5q, q, 0.5q, 1]$, so the number of nodes in the first hidden layer is larger than the number of features, and with a high probability the DNN is able to capture the information of the features and transfer them through the network. 
	Setting the number of nodes in the first hidden layer smaller than that in the input layer may result in losing some input information. 
	
	We continue training until we meet one of the following criteria:
	\begin{itemize}
		\item the point-wise improvement in loss function value is less than $0.01\%$, or
		\item the number of passes over the training data reaches {\tt MaxEpoch}.
	\end{itemize}
	(We set {\tt MaxEpoch}=100.)
	
	Of course, we cannot guarantee that this approach will produce an optimal network structure, but it eliminates the work of determining the structure, and our experiments suggest that it performs well. We also note that one still must follow an approach to determine a suitable learning rate and regularization parameter (see \cite{snoek2012practical, eggensperger2013towards, domhan2015speeding, bergstra2012random}).

	In order to see how well the fixed-size network works, we ran the same experiments as in Section \ref{sec:Numerical_experiments_simulation}. 
	In these tests, we fixed the network structure to $[q, 1.5q, q, 0.5q, 1]$ with learning rate $=0.001$ and $\lambda=0.005$ for all demand distributions. 
	In all cases except normally distributed demand, the network obtained near-optimal costs after at most 10 epochs \add{(which, on average, took 10 minutes to train)}, when improvement stopped. 
	For normally distributed demands, the algorithm ran for at least $50$ epochs to get a converged network.	
	Table \ref{tb:fixed_str_comparison_100_200_epochs} shows the results of the test datasets for all demand distributions, in which we provide the gap between the results of the fixed network and the results from the HyperBand algorithm. 
	As provided in the table, when we train for 100 epochs, the fixed network obtains costs that are very close to those obtained using the HyperBand algorithm. 
	For 1, 10, 100, and 200 clusters, it obtains average gaps of $0.67\%$, $1.9\%$, $43.5\%$, and $65.3\%$ compared to the results of networks obtained by HyperBand algorithm.
\add{

	\linespread{1}

	\begin{table}[]
		\centering
		\caption{Results of 100 and 200 training epochs.}
		\label{tb:fixed_str_comparison_100_200_epochs}
		\begin{tabular}{lcccc|cccc}
			& \multicolumn{4}{c}{100 epochs}            & \multicolumn{4}{c}{300 epochs}          \\ \cline{2-9} 
			\multicolumn{1}{l|}{clusters} & 1           & 10          & 100         & 200         & 1          & 10          & 100         & 200         \\ \hline
			\multicolumn{1}{l|}{normal}      & 0.000       & 0.004           & 2.006       & 3.083       & 0.000          & 0.004       & 0.005       & 3.083       \\
			\multicolumn{1}{l|}{lognormal}   & 0.003       & 0.006           & 0.129       & 0.006       & 0.000          & 0.004       & 0.126       & 0.011       \\
			\multicolumn{1}{l|}{uniform}     & 0.001       & 0.012           & 0.020       & 0.134       & 0.000          & 0.001       & 0.020       & -0.004      \\
			\multicolumn{1}{l|}{beta}        & 0.029       & 0.003           & 0.014       & 0.023       & 0.027          & -0.006      & 0.007       & 0.021       \\
			\multicolumn{1}{l|}{exponential} & 0.000       & 0.071           & 0.008       & 0.019       & 0.000          & 0.001       & 0.006       & 0.018       \\ \hline
			\multicolumn{1}{l|}{average}     & 0.0067 & 0.0192 & 0.4353 & 0.6531 & 0.0054 & 0.0008 & 0.0329 & 0.6260 \\ \hline
		\end{tabular}
	\end{table}

	\linespread{2}
}

	In order to see the effect of training length, we ran all experiments for 300 epochs to see whether the solutions improve, which are provided in right side of Table \ref{tb:fixed_str_comparison_100_200_epochs}. 
	The average gaps decreased to $0.5\%$, $0.08\%$, $3.29\%$, and $62.6\%$ for 1, 10, 100, and 200 clusters, respectively. 
	Therefore, running the DNN for longer training periods can help to get smaller cost values. 

	In sum, setting the network size using this approach is much cheaper than any extension of random search or Bayesian optimization, and it can provide near-optimal results for the newsvendor problem when there is a sufficiently large number of historical records. (In our experiment, this corresponds to having fewer clusters.) When there is insufficient historical data available, additional tuning and/or training is required in order to obtain good results.

\add{

} 

\end{document}